\documentclass[fleqn,10pt]{wlscirep}
\usepackage[utf8]{inputenc}
\usepackage[T1]{fontenc}
\usepackage{multirow}
\usepackage{placeins}
\usepackage{comment}
\usepackage{textcomp}

\title{Toward a universal foundation model for graph-structured data}

\author[1]{Sakib Mostafa}
\author[1,2,3,*]{Lei Xing}
\author[1,*]{Md Tauhidul Islam}
\affil[1]{Department of Radiation Oncology, Stanford University, Stanford, CA, USA}
\affil[2]{Institute of Computational and Mathematical Engineering, Stanford University, Stanford, CA, USA}
\affil[3]{Department of Electrical Engineering, Stanford University, Stanford, CA, USA}

\affil[*]{Correspondence: tauhid@stanford.edu \& lei@stanford.edu}

\begin{abstract}
Graphs are a central representation in biomedical research, capturing molecular interaction networks, gene regulatory circuits, cell--cell communication maps, and knowledge graphs. Despite their importance, currently there is not a broadly reusable foundation model available for graph analysis comparable to those that have transformed language and vision. Existing graph neural networks are typically trained on a single dataset and learn representations specific only to that graph's node features, topology, and label space, limiting their ability to transfer across domains. This lack of generalization is particularly problematic in biology and medicine, where networks vary substantially across cohorts, assays, and institutions. Here we introduce a graph foundation model designed to learn transferable structural representations that are not specific to specific node identities or feature schemes. Our approach leverages feature-agnostic graph properties, including degree statistics, centrality measures, community structure indicators, and diffusion-based signatures, and encodes them as structural prompts. These prompts are integrated with a message-passing backbone to embed diverse graphs into a shared representation space. The model is pretrained once on heterogeneous graphs and subsequently reused on unseen datasets with minimal adaptation. Across multiple benchmarks, our pretrained model matches or exceeds strong supervised baselines while demonstrating superior zero-shot and few-shot generalization on held-out graphs. On the SagePPI benchmark, supervised fine-tuning of the pretrained backbone achieves a mean ROC-AUC of 95.5\%, a gain of 21.8\% over the best supervised message-passing baseline. The proposed technique thus provides a unique approach toward reusable, foundation-scale models for graph-structured data in biomedical and network science applications.
\end{abstract}

\begin{document}

\flushbottom
\maketitle
\thispagestyle{empty}

\section{Introduction}

\begin{figure}[pth]
\centering
\includegraphics[width=0.8\linewidth, page=1]{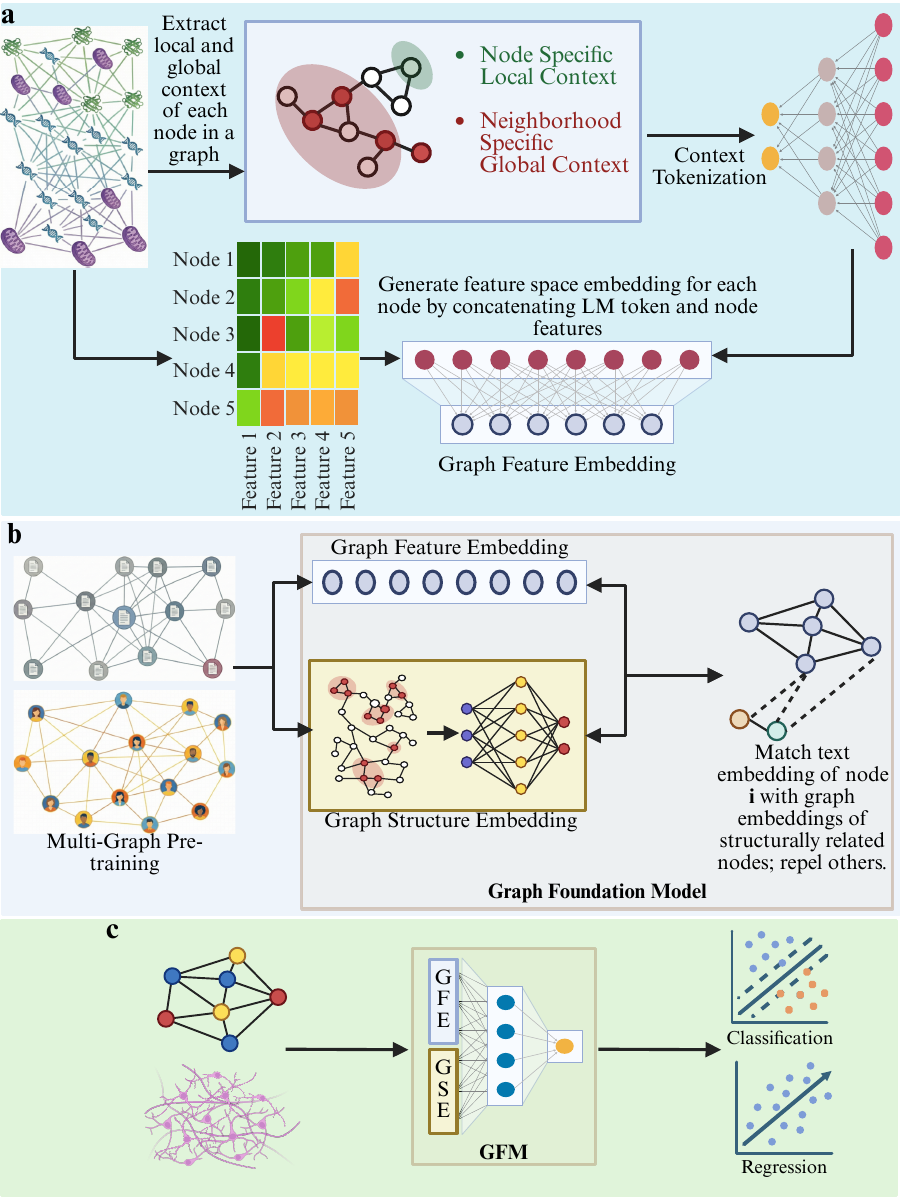}
\caption{\textbf{Architecture of the Graph Foundation Model.}
\textbf{a}, Structural prompt construction. For each node, local
context descriptors---degree, clustering coefficient, $k$-core
number, ego-network statistics, and PageRank---are combined with
global context indicators derived from two community detection
algorithms, Label Propagation and SCoDA, each contributing
community identity, community size, and intra-community density.
These descriptors are assembled into a natural-language node
profile and tokenized by a miniature language model (MiniLM) to
produce a context embedding. Node features are projected into a
matching space and concatenated with this embedding to form the
Graph Feature Embedding (GFE), a representation that encodes both
attribute content and structural role without relying on
dataset-specific feature semantics.
\textbf{b}, Multi-graph pretraining. A GraphSAGE backbone ingests
multiple heterogeneous graphs simultaneously, producing a Graph
Structure Embedding (GSE) for every node via neighborhood
aggregation. Contrastive alignment is applied between the GFE and
the GSE: the text embedding of node $i$ is matched against the
graph embeddings of structurally related nodes and repelled from
structurally dissimilar nodes, training the backbone to map
diverse graphs into a shared topological representation space.
\textbf{c}, Downstream deployment. The frozen or fine-tuned GFM
generates GFE and GSE representations for nodes in an unseen
graph. These are concatenated and passed to a lightweight
task-specific head for node-level classification or regression,
requiring minimal labeled data on the target graph.}
\label{fig:fig1_pipeline}
\end{figure}

Graphs are widely used to represent relational systems in biomedical science, such as molecular interaction networks~\cite{barabasi2004network}, gene regulatory circuits~\cite{davidson2002genomic}, and cell--cell communication maps~\cite{armingol2021cell}. They are also the inherent representation for biomedical knowledge graphs that link genes, diseases, drugs, and phenotypes~\cite{nicholson2020constructing}. This wide applicability of graph representations makes graph learning critically important for both biological discovery and clinical translation. In recent years, deep learning has transformed representation learning in domains with fixed input formats, most notably language and vision, where large pretrained models can be reused across tasks with minimal adaptation~\cite{brown2020language,dosovitskiy2021image,bommasani2021foundation}. However, despite major advances in graph neural networks, there is still no broadly reusable foundation model that generalizes reliably across graphs from different domains. This challenge is particularly pronounced in biology and medicine, where graph structure can vary across cohorts, assays, and institutions, causing models trained on one graph to transfer poorly to another.

Most existing graph neural networks are trained on a single dataset and learn representations that are specific to that graph's node features, topology, and label space~\cite{wu2021survey}. Common message-passing architectures, including Graph Convolutional Networks (GCN)~\cite{kipf2017gcn}, Graph Isomorphism Networks (GIN)~\cite{xu2019gin}, Graph Attention Networks (GAT)~\cite{velickovic2018gat}, and GraphSAGE~\cite{hamilton2017inductive}, can be used on semi-supervised learning and node classification tasks when training and testing occur within the same graph. However, the learned parameters typically reflect the statistics and semantics of one dataset at a time. Recent self-supervised approaches, including Deep Graph Infomax (DGI)~\cite{velickovic2019dgi}, GRACE~\cite{zhu2020grace}, and BGRL~\cite{thakoor2022bgrl}, have the same limitations. They define objectives on a single graph and learn representations that are most reliable within that same structural and feature distribution. As a result, graph representation learning remains dataset-specific in practice. Models are often retrained from scratch for each new graph, and knowledge learned from one dataset is not consistently carried forward to another~\cite{zhu2020transfer}, even when the graphs describe related biological systems.

This generalizability limitation stems from properties that are inherent to graphs. Unlike images or text, graphs do not have a canonical input space~\cite{bronstein2021geometric}. Node features can take many forms, such as bag-of-words vectors, molecular fingerprints, continuous attributes, or learned embeddings, and their meaning varies widely across applications. The meaning of edges also changes across domains. An edge may represent a physical interaction, a regulatory relationship, a citation, or a transactional link. This variability makes it difficult to define a shared representation that works across datasets. There are also architectural constraints. Most graph models rely on local message passing, where each layer aggregates information from a node's neighbors. This works well for local structure, but long-range dependencies require deeper networks. As depth increases, node representations can become overly similar~\cite{li2018oversmoothing}, and information from distant regions of the graph can be compressed into small vectors. These effects weaken the model's ability to preserve global structural signals consistently. Pretraining objectives do not automatically fix this. Many objectives are tuned to the distribution of a single graph and may not remain stable under large shifts in structure, feature space, or semantics. Together, these issues make cross-graph generalization difficult, even when models perform well within a single dataset.

For a reusable foundation model, the ideal representation must rely on components of graph organization that are intrinsic and broadly available, rather than being tied to dataset-specific features or labels. Graph topology provides such a basis. Many structural properties are defined without reference to feature semantics. These include degree statistics, centrality measures, community structure, diffusion statistics, and higher-order connectivity patterns~\cite{newman2003structure}. They can be computed for any graph and can be compared across graphs, even when the underlying node attributes differ. If these structural signals can be expressed through a common interface, then heterogeneous graphs may be mapped into a shared embedding space without requiring common node identities or a shared feature schema. This provides a practical route to cross-domain learning in settings where raw features are incompatible, but topology remains meaningful.

Here, to address the limitations of existing graph analysis methods and utilize the generalizable graph structural properties, we introduce a graph foundation model that learns transferable structural representations not tied to any single graph's node identities or feature schema. Our approach converts feature-agnostic graph properties into structured prompts at the node level. These prompts summarize a node's position and role in the graph using degree statistics, centrality measures, community indicators, and diffusion-based signatures, along with related structural descriptors such as clustering and core structure. When textual node descriptors are available, they are included as additional context. We integrate these prompts with a lightweight message-passing backbone. The prompts provide an explicit structural description, while message passing captures local relational patterns that depend on connectivity. Training aligns these two signals so that the backbone learns representations that remain meaningful across graphs with different feature spaces. Because the structural interface is defined from topology, the model can align graphs without requiring common node identities, shared attributes, or shared label sets, which are rarely available across biomedical datasets.

We pretrain the model once on a heterogeneous collection of graphs and then reuse the frozen backbone on unseen datasets for different downstream applications. We evaluate our approach on four held-out biological network benchmarks: SagePPI, a human protein--protein interaction network with GO biological process annotations~\cite{hamilton2017inductive}; ogbn-proteins, a large-scale STRING-derived interaction network from the Open Graph Benchmark with species-stratified evaluation~\cite{hu2020ogb,szklarczyk2019string}; StringGO, a STRING network with GO annotations spanning three ontologies evaluated using amino acid composition features~\cite{szklarczyk2019string,go2021}; and Fold-PPI, a collection of 144 tissue-specific interaction networks with protein structural fold class labels drawn from SCOP~\cite{huang2020gmeta,murzin1995scop}. Across these benchmarks, our pretrained model matches or exceeds strong supervised baselines in zero-shot mode and shows substantially improved generalization when labels are limited. These findings indicate that pretraining on heterogeneous graphs produces representations that capture biologically meaningful organization without requiring task-specific training on the target network.

\section{Results}
\subsection{Topology-based pretraining produces transferable functional representations in a human protein interaction network}

\begin{figure}[pth]
\centering
\includegraphics[width=0.8\linewidth, page=2]{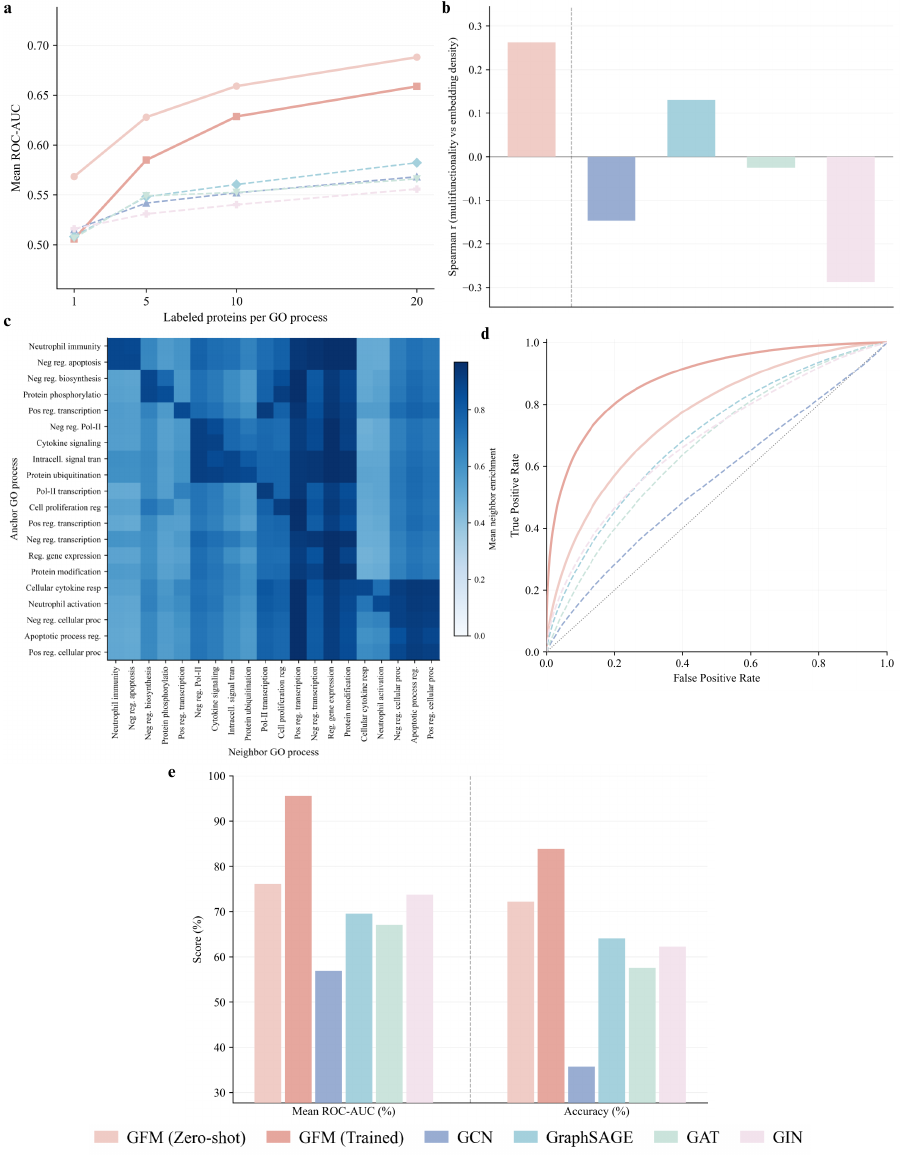}
\caption{\textbf{The pretrained GFM encodes protein functional organization
in SagePPI and generalizes with minimal supervision.}
\textbf{a}, Few-shot generalization curves. Mean ROC-AUC is shown as a
function of the number of labeled proteins per GO process, evaluated on
held-out test proteins using a logistic probe trained on K positive and K
negative examples per label (K $\in$ \{1, 5, 10, 20\}). Solid lines denote
GFM variants; dashed lines denote supervised baselines.
\textbf{b}, Multifunctionality--embedding density correspondence. Bars show
the Spearman rank correlation between the number of GO process annotations
per protein and the local embedding density computed from the 15 nearest
neighbors in each model's embedding space. The dashed vertical line separates
the pretrained GFM zero-shot model (left) from supervised GNN baselines
(right). Positive values indicate that highly annotated proteins occupy denser
embedding neighborhoods.
\textbf{c}, Cross-functional co-enrichment in GFM Trained embeddings. Each
cell reports the mean fraction of GO-annotated neighbors that a protein from
one anchor process accumulates within a 25-nearest-neighbor embedding
neighborhood annotated by another process. Rows and columns are ordered by
hierarchical clustering of the symmetrized co-enrichment matrix. Darker
entries indicate higher cross-process neighborhood sharing.
\textbf{d}, Macro-averaged ROC curves across all 121 GO process labels.
Each curve corresponds to the mean interpolated true positive rate at fixed
false positive rate thresholds, averaged across labels with at least two
represented classes. The dotted diagonal denotes random performance.
\textbf{e}, Summary classification performance. Mean ROC-AUC and accuracy
are shown for GFM zero-shot, GFM Trained, and four supervised GNN baselines
(GCN, GraphSAGE, GAT, GIN). The dashed vertical line separates the two
performance metrics. All models use the same train/validation/test split.}
\label{fig:sageppi}
\end{figure}

We first evaluated the GFM on the SagePPI protein--protein interaction network, which comprises experimentally supported physical interactions among human proteins annotated with 121 Gene Ontology (GO)~\cite{ashburner2000go}
biological process labels drawn from the MSigDB GO:BP 2021
collection~\cite{liberzon2015msigdb}. Each protein carries multiple functional annotations, making this a multilabel node classification task that simultaneously probes whether learned embeddings organize the interaction network according to coherent biological programs. SagePPI represents a stringent generalization benchmark: the model was pretrained on a heterogeneous collection of graphs that does not include this network, and evaluation is performed on proteins held out during pretraining.

The pretrained GFM in zero-shot mode, using a frozen backbone with a lightweight linear probe, achieved a mean ROC-AUC of 76.1\% and an accuracy of 72.2\% across all 121 GO process labels (Fig.~\ref{fig:sageppi}e). This exceeded the strongest supervised baseline, GIN~\cite{xu2019gin}, which reached a mean ROC-AUC of 73.7\% under the same evaluation protocol despite being trained directly on SagePPI node features and topology. GraphSAGE~\cite{hamilton2017inductive} and
GAT~\cite{velickovic2018gat} reached 69.7\% and 67.1\%
respectively, while GCN~\cite{kipf2017gcn} achieved 57.1\%. When the GFM backbone was fine-tuned on SagePPI with full supervision, mean ROC-AUC increased substantially to 95.5\% and accuracy to 84.1\%, representing a gain of 21.8 percentage points over the best supervised baseline. Macro-averaged ROC curves across all 121 labels confirm that both GFM variants dominate the baselines at every operating threshold, with the separation most
pronounced at low false positive rates where precision requirements are most stringent (Fig.~\ref{fig:sageppi}d; Supplementary Fig.~\ref{fig:roc_full}).

To assess whether this advantage reflects a general capacity for rapid
adaptation rather than performance that requires full supervision, we
evaluated each model under a few-shot protocol in which a logistic probe
was trained using only $K$ labeled examples per GO process, with $K
\in \{1, 5, 10, 20\}$ (Fig.~\ref{fig:sageppi}a). At $K = 1$, the GFM
zero-shot embeddings achieved a mean ROC-AUC of 0.567, while all
supervised baselines clustered between 0.505 and 0.515 at this extreme
label scarcity. At $K = 20$, GFM zero-shot reached 0.690 and GFM
Trained reached 0.661, while the best-performing baseline, GraphSAGE,
reached only 0.583. The pretrained GFM thus achieves at $K = 1$ a level
of performance that supervised baselines approach only after accumulating
20 labeled examples per GO process. GFM Trained embeddings showed a
complementary trajectory: starting near baseline level at $K = 1$ and
rising steeply as labels accumulate, consistent with the interpretation
that fine-tuned representations encode more discriminative task-specific
structure that benefits substantially from downstream supervision.
Across the entire few-shot range, both GFM variants outperformed all
supervised baselines, indicating that pretraining instills transferable
functional structure that task-specific training from scratch does not
recover even with equivalent labeled data.

Beyond discriminative performance, we examined whether GFM embeddings
organize proteins according to biologically coherent functional
neighborhoods. For six representative GO processes spanning distinct
biological programs---DNA repair, cell cycle regulation, Pol-II
transcription, cell proliferation regulation, negative regulation of
transcription, and nervous system development---we computed the
same-label fraction among the $k$ nearest neighbors in each model's
embedding space for $k \in \{5, 10, 20, 30, 50, 75, 100\}$
(Supplementary Figs.~S\ref{fig:enrichment_s2},
S\ref{fig:enrichment_s3}). GFM zero-shot and GFM Trained embeddings
consistently maintained higher same-label fractions at all values of $k$
relative to all baselines, with the gap most pronounced at small $k$
where local neighborhood purity is most informative about embedding
organization. This advantage was reproducible across all six processes
without exception, indicating that functional cohesion in the GFM
embedding space is not confined to a single annotation class but
reflects a broadly organized representation of protein function.

To characterize the joint functional structure of the embedding space,
we computed a co-enrichment matrix across the 20 most populated GO
processes, quantifying the mean fraction of annotated neighbors that a
protein from one process accumulates within a 25-nearest-neighbor
embedding neighborhood defined by GFM Trained representations
(Fig.~\ref{fig:sageppi}c). Hierarchical clustering of this matrix
revealed a block structure in which GO processes with established
biological relationships clustered together without prior knowledge of
their groupings. Transcription-related processes---Pol-II transcription,
positive and negative regulation of transcription, and regulation of
gene expression---formed a coherent cluster with elevated inter-process
enrichment, consistent with their shared mechanistic basis in gene
regulatory circuits. Immune and cytokine-related processes---neutrophil
immunity, neutrophil activation, cytokine signaling, and cellular
cytokine response---formed a separate block, reflecting their shared
involvement in inflammatory programs. These patterns indicate that the
GFM embedding space preserves not only within-class functional cohesion
but also the higher-order relational structure among biological
processes, a property that emerges from topology-based pretraining
rather than from any explicit annotation of process relationships.

We further examined whether the embedding density of a protein in the
GFM zero-shot space reflects its degree of multifunctionality, defined
as the number of GO process annotations assigned to that protein
(Fig.~\ref{fig:sageppi}b). Spearman rank correlation between per-protein
label count and local embedding density yielded $r = +0.263$ for GFM
zero-shot. In contrast, GCN produced $r = -0.131$, GraphSAGE $r =
+0.130$, GAT $r = -0.025$, and GIN $r = -0.290$. The strongly negative
correlation in GIN embeddings indicates that supervised training under
this architecture places highly multifunctional proteins at the
periphery of the embedding space, which is inconsistent with what
protein interaction network topology would predict. The positive
correlation in GFM zero-shot embeddings is consistent with the
expectation that proteins participating in many biological processes
occupy structurally central positions in the interaction network~\cite{jeong2001lethality} and
should therefore be embedded in densely populated regions of a
topology-informed representation space. This property emerges without
any supervision from functional labels and therefore reflects structure
that pretraining captures directly from graph topology.

Taken together, these results establish that the pretrained GFM encodes
transferable structural representations that organize the SagePPI
interaction network according to protein functional programs. In
zero-shot mode, the frozen backbone with a linear probe surpasses the
best supervised GNN baseline, and in the few-shot regime it achieves
with a single labeled example per process what supervised models require
20 labeled examples to approach. The embedding space preserves
functional cohesion across diverse GO biological processes, recovers
known relationships among annotation classes through its co-enrichment
structure, and places multifunctional proteins in structurally central
positions. These findings indicate that structural pretraining encodes
functional organization at a level that does not depend on the
specific proteins, species, or annotation vocabulary present
during training. Both benchmarks evaluated so far use node features derived from
interaction databases and expression data. A separate question is
whether the same structural representations transfer to a graph
whose node features are qualitatively different, specifically
primary sequence composition rather than pathway or expression
signals.

\subsection{Cross-species generalization scales with GO annotation specificity in ogbn-proteins}

\begin{figure}[pth]
\centering
\includegraphics[width=0.85\linewidth, page = 3]{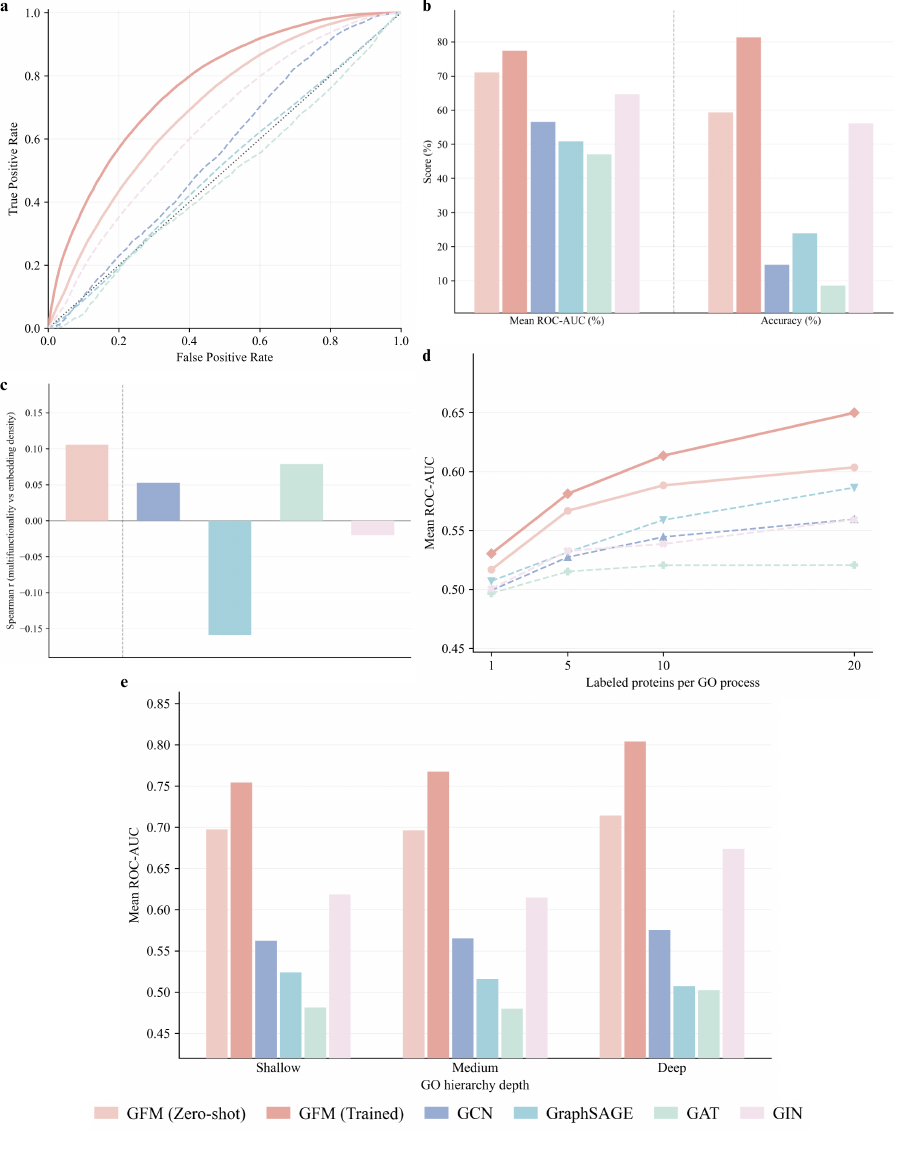}
\caption{\textbf{The pretrained GFM generalizes across the ogbn-proteins
benchmark and captures GO hierarchy structure.}
\textbf{a}, Macro-averaged ROC curves across all GO term labels for
GFM zero-shot, GFM Trained, and four supervised GNN baselines (GCN,
GraphSAGE, GAT, GIN). Solid lines denote GFM variants; dashed lines
denote supervised baselines. The dotted diagonal denotes random
performance.
\textbf{b}, Summary classification performance. Mean ROC-AUC and
accuracy are shown for all six models. The dashed vertical line
separates the two reported metrics. All models use the standard
ogbn-proteins train/validation/test split.
\textbf{c}, Multifunctionality--embedding density correspondence.
Bars show the Spearman rank correlation between the number of GO
annotations per protein and local embedding density computed from
the 15 nearest neighbors. The dashed vertical line separates the
pretrained GFM zero-shot model from supervised baselines. Positive
values indicate that highly annotated proteins occupy denser
embedding neighborhoods.
\textbf{d}, Few-shot generalization curves. Mean ROC-AUC is shown
as a function of the number of labeled proteins per GO process
($K \in \{1, 5, 10, 20\}$), evaluated on held-out test proteins
using a logistic probe trained on $K$ positive and $K$ negative
examples per label. Solid lines denote GFM variants; dashed lines
denote supervised baselines.
\textbf{e}, GO hierarchy depth stratification. Mean ROC-AUC is
shown separately for GO terms at shallow, medium, and deep levels
of the GO biological process hierarchy. GFM Trained performance
increases with hierarchy depth while baseline performance remains
flat or declines.}
\label{fig:ogbnproteins}
\end{figure}

We next evaluated the GFM on the ogbn-proteins benchmark from the Open Graph Benchmark~\cite{hu2020ogb}, a large-scale protein--protein interaction network derived from the STRING database~\cite{szklarczyk2019string} with 112 GO biological process annotations per node as multilabel targets. The benchmark uses a species-stratified split in which test proteins come from species not represented during training. This makes cross-species generalization a requirement for strong test performance, not an optional property. A model that memorizes the training graph will fail on the test set. We used this to ask directly whether structural pretraining produces embeddings that transfer across biological contexts.

The pretrained GFM in zero-shot mode achieved a mean ROC-AUC of
71.0\% across all 112 GO labels (Fig.~\ref{fig:ogbnproteins}b).
This exceeded GCN (56.5\%), GraphSAGE (50.9\%), and GAT (47.0\%),
and approached GIN (64.6\%), the strongest supervised baseline.
What makes this margin meaningful is that GIN operates on
confidence-weighted STRING edges during training~\cite{szklarczyk2019string}, providing explicit
interaction reliability information that the GFM never receives.
When the GFM backbone was fine-tuned with full supervision, mean
ROC-AUC increased to 77.5\% with an accuracy of 81.1\%, a gain of
12.9 percentage points over GIN. Macro-averaged ROC curves confirm
that both GFM variants dominate the baselines at every operating
threshold (Fig.~\ref{fig:ogbnproteins}a;
Supplementary Fig.~\ref{fig:ogbn_roc_full}).

The few-shot evaluation reinforces these findings
(Fig.~\ref{fig:ogbnproteins}d). At $K = 1$ labeled protein per GO
process, GFM zero-shot achieved a mean ROC-AUC of 0.620. All
supervised baselines fell between 0.495 and 0.540 at this shot
count despite having been trained on the full graph with complete
label supervision. At $K = 20$, GFM zero-shot reached 0.738 while
the best baseline reached 0.646. The GFM few-shot curve rises
steeply from $K = 1$ and continues improving, indicating that
pretrained structural embeddings provide a strong initialization
that benefits from small amounts of additional supervision. Baseline
trajectories remain comparatively flat across the same range.

The most distinctive result in this dataset concerns GO hierarchy
depth (Fig.~\ref{fig:ogbnproteins}e). Shallow GO terms are broad
and annotate many proteins~\cite{go2021}. Deep terms are specific and annotate
far fewer. For GFM Trained, mean ROC-AUC increased monotonically
from 75.4\% at shallow terms to 76.7\% at medium depth and 80.5\% at
deep terms. GIN showed no consistent directional trend, producing
61.7\% at shallow, 61.4\% at medium, and 67.5\% at deep. GFM zero-shot
similarly improved with depth, from 70.0\% at shallow to 71.4\% at
deep. The GFM widens its advantage over baselines precisely where
the task is hardest. Deep GO terms require integrating information
across long network paths and community boundaries to identify
specific functional roles. Structural pretraining captures these
signals through diffusion statistics, $k$-core
decomposition~\cite{batagelj2003core}, and
multi-scale community indicators~\cite{hollocou2017scoda,raghavan2007near}. Supervised models trained on a
single graph do not acquire this capacity from labeled data alone.

The multifunctionality analysis replicates the SagePPI result on a
graph with a different annotation system and edge weighting scheme
(Fig.~\ref{fig:ogbnproteins}c). GFM zero-shot produced a Spearman
rank correlation of $r = +0.105$ between per-protein GO annotation
count and local embedding density. GCN produced $r = +0.052$,
GAT $r = +0.079$, GraphSAGE $r = -0.155$, and GIN $r = -0.020$.
The positive correlation appears in GFM embeddings on both SagePPI
and ogbn-proteins, across different databases and annotation systems.
This consistency supports the interpretation that pretraining from
topology places structurally central, highly annotated proteins in
dense embedding regions as a general property of the learned
representation space.

Co-enrichment analysis of the GFM Trained embedding space shows
that biological processes are organized into functionally coherent
blocks without any explicit knowledge of process relationships
(Supplementary Fig.~\ref{fig:ogbn_coinrich}). An immune module
comprising regulation of immune system process, antigen processing
and presentation, acute inflammatory response, immune system process,
and immune effector process shows enrichment ratios substantially
above 1.0 within the block. A cytokine signaling module covering
positive regulation of cytokine production, regulation of cytokine
production, cytokine production, and cell activation shows strong
internal enrichment and cross-enrichment with the immune block.
This is consistent with the known coupling between cytokine-mediated
signaling and acute inflammatory programs~\cite{dinarello2000cytokines}. Same-label enrichment
curves across six GO annotation categories confirm that GFM
embeddings maintain higher functional neighborhood purity at all
values of $k$, with the advantage widening for the more specific
metabolic process terms relative to broad categories such as
molecular function (Supplementary Fig.~\ref{fig:ogbn_enrichment}).

The ogbn-proteins results strengthen the core claim of this work
in a specific way. The species-stratified split means that the test
proteins come from biological contexts the model has never seen.
The GFM zero-shot model still exceeds three supervised baselines
and narrows the gap with GIN to 6.4 percentage points under these
conditions. Its advantage over baselines grows with GO hierarchy
depth rather than shrinking, which is the opposite of what a model
relying on local pattern memorization would produce. These findings
indicate that structural pretraining encodes functional organization
at a level that does not depend on the specific proteins, species,
or annotation vocabulary present during training.

\subsection{Structural pretraining recovers subcellular compartment organization from amino acid sequence networks in StringGO}

\begin{figure}[pth]
\centering
\includegraphics[width=0.95\linewidth, page = 4]{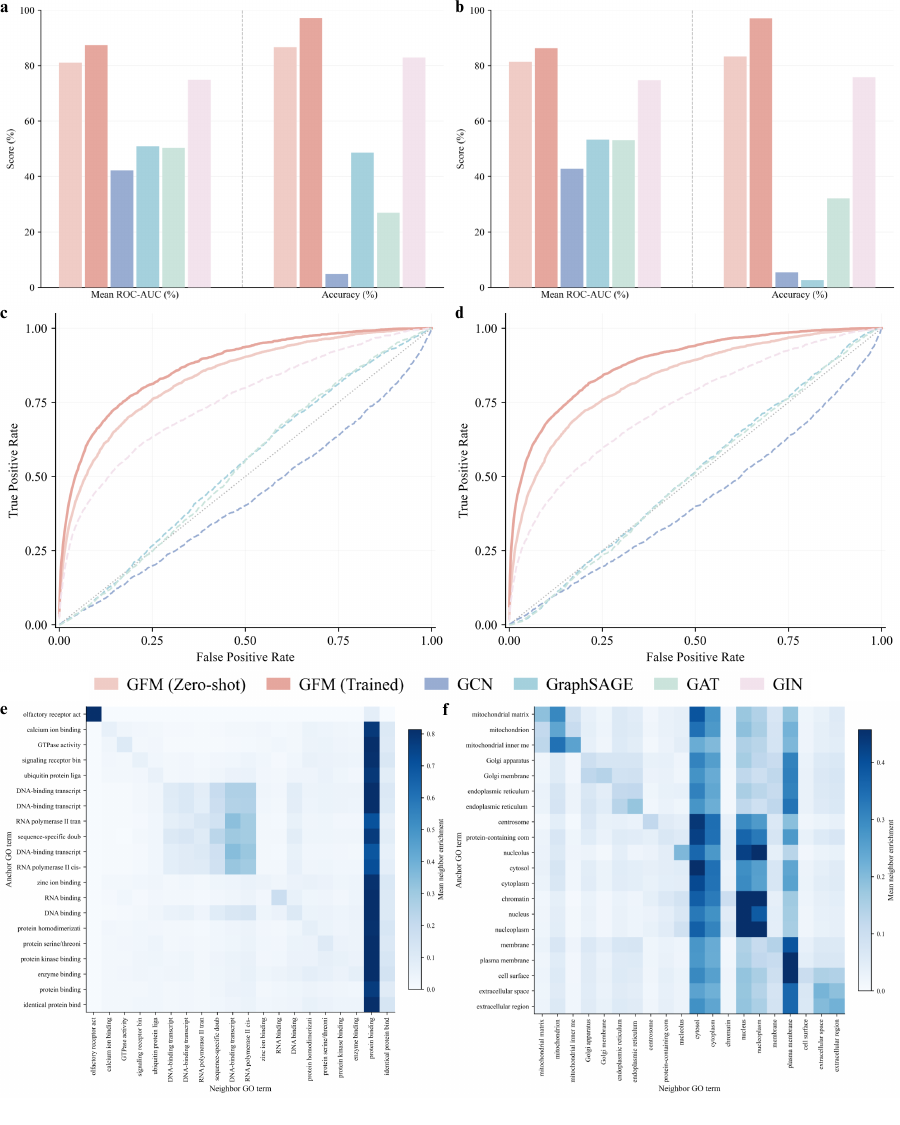}
\caption{\textbf{The pretrained GFM generalizes across GO ontologies
in the STRING interaction network.}
\textbf{a}, Mean ROC-AUC and accuracy for Molecular Function (MF)
GO term prediction. All six models are compared under the same
train/validation/test split on the StringGO network.
\textbf{b}, Mean ROC-AUC and accuracy for Cellular Component (CC)
GO term prediction under identical conditions.
\textbf{c}, Macro-averaged ROC curves for MF. Solid lines denote
GFM variants; dashed lines denote supervised baselines. The dotted
diagonal denotes random performance.
\textbf{d}, Macro-averaged ROC curves for CC. Same protocol as
\textbf{c}.
\textbf{e}, Cross-functional co-enrichment in GFM Trained
embeddings for MF. Each cell reports the mean fraction of GO-term
annotated neighbors that a protein from one anchor MF term
accumulates within a 25-nearest-neighbor embedding neighborhood
annotated by another term. Rows and columns are ordered by
hierarchical clustering of the symmetrized co-enrichment matrix.
\textbf{f}, Cross-functional co-enrichment for CC. Same protocol
as \textbf{e}. Three distinct spatial modules are visible:
a mitochondrial module, a nuclear and cytoplasmic module, and
a membrane module.}
\label{fig:stringgo}
\end{figure}

SagePPI and ogbn-proteins use node features drawn from interaction
and expression databases that share at least partial conceptual
overlap with signals the model may have encountered during
pretraining. StringGO removes this overlap. It is a human protein
interaction network derived from the STRING
database~\cite{szklarczyk2019string} with GO annotations evaluated
separately across three ontologies: Molecular Function (MF),
Biological Process (BP), and Cellular Component (CC)~\cite{go2021}.
MF annotations describe the biochemical activities of individual
proteins. BP annotations describe the larger biological programs
in which proteins participate. CC annotations describe the
subcellular locations where proteins are active. Each ontology
defines an independent multilabel classification task on the same
underlying interaction graph. Node features are 20-dimensional
amino acid composition vectors encoding local sequence triplet
statistics, which carry primary sequence properties rather than
expression or pathway information. This setting tests whether
structural pretraining generalizes to a graph whose node features
carry a fundamentally different biological meaning from those seen
during training.

Across MF and CC, the pretrained GFM in zero-shot mode achieved
mean ROC-AUC values of 81.4\% and 81.1\% respectively, exceeding
the strongest supervised baseline, GIN, which reached 74.7\% on
MF and 74.9\% on CC (Fig.~\ref{fig:stringgo}a,b). GCN, GraphSAGE,
and GAT fell substantially further behind, reaching between 42.2\%
and 53.4\% across the two ontologies. When the GFM backbone was
fine-tuned with full supervision, mean ROC-AUC increased to 86.3\%
on MF and 87.4\% on CC, representing gains of 11.6 and 12.5
percentage points respectively over GIN. Macro-averaged ROC curves
confirm the consistent separation between GFM variants and all
baselines across the full threshold range for both MF and CC
(Fig.~\ref{fig:stringgo}c,d). The margin over GCN, GraphSAGE, and
GAT is particularly large, exceeding 30\% in
mean ROC-AUC, which indicates that standard message-passing
architectures struggle to extract functional signal from amino
acid composition features alone without the structural scaffolding
that pretraining provides.

On the BP ontology, the GFM zero-shot model achieved a mean
ROC-AUC of 80.5\%, which again exceeded all supervised baselines
including GIN (69.3\%), GCN (44.9\%), GraphSAGE (50.6\%), and
GAT (51.4\%) (Supplementary Fig.~\ref{fig:stringgo_bp}).
The zero-shot advantage on BP is therefore consistent with MF
and CC. However, the GFM Trained model reached only 68.7\% on
BP, which did not improve over the zero-shot baseline and was
marginally below GIN. This behavior is specific to the BP
fine-tuning and is not observed in the other two ontologies.
The BP label space is substantially larger and more heterogeneous
than MF or CC, and fine-tuning on it is more prone to instability
under the current training configuration. The zero-shot result
nonetheless demonstrates that the pretrained embeddings encode
BP-relevant functional organization, and the failure of supervised
fine-tuning on BP does not undermine the generalization claim.

The few-shot protocol used here trains a logistic probe on $K$
proteins per GO term drawn from the training split, then evaluates
on all remaining proteins. At $K = 1$, GFM zero-shot achieved
ROC-AUC of 0.657, 0.692, and 0.666 on MF, BP, and CC respectively.
At $K = 20$, these values reached 0.810, 0.804, and 0.802. The
consistent improvement across all three ontologies indicates that
the pretrained embeddings remain informative across very different
types of GO annotation, from the specific biochemical activities
captured by MF to the broad cellular processes captured by BP.
Baseline few-shot results are not reported here because the
few-shot protocol was designed to evaluate embedding quality rather
than retraining capacity, and the baselines produce embeddings that
are trained on a specific label set and do not generalize in the
same way.

GFM embeddings also retain functional structure under amino acid
feature perturbation. At 50\% noise applied to the 20-dimensional
amino acid composition input, GFM zero-shot maintained a mean
ROC-AUC of 0.791 on MF and 0.791 on CC, representing a decline
of only 2.3 and 2.0 percentage points from the clean-feature
baseline respectively
(Supplementary Fig.~\ref{fig:stringgo_noise}). This stability
reflects the fact that the GFM embedding is primarily driven by
the structural topology tokens, which encode community membership,
diffusion statistics, and local connectivity and are not affected
by amino acid feature perturbation.

Co-enrichment analysis of the GFM Trained embedding space reveals
biologically coherent term groupings for both MF and CC
(Fig.~\ref{fig:stringgo}e,f). In the MF heatmap, a cluster of
DNA-binding and RNA polymerase II activity terms shows elevated
mutual enrichment, consistent with the shared mechanistic role
of these factors in transcription initiation and elongation~\cite{cramer2019rnapol}. The
olfactory receptor activity term shows strong self-enrichment and
high enrichment with the broader protein binding category,
reflecting the well-known protein-protein interaction network
of olfactory receptors and their shared structural scaffold~\cite{buck1991olfactory}.
The CC heatmap shows three clearly resolved modules. A
mitochondrial module groups mitochondrial matrix, mitochondrion,
and mitochondrial inner membrane together with high mutual
enrichment. A nuclear and cytoplasmic module groups nucleus,
nucleoplasm, cytosol, cytoplasm, and chromatin, capturing the
shared spatial occupancy of nuclear proteins within the cell.
A membrane module groups plasma membrane and membrane, which
is consistent with their overlapping protein populations. These
three modules correspond directly to the major spatial
compartments of eukaryotic cells and emerge from network topology
alone without any explicit annotation of compartment membership
provided to the model. The BP co-enrichment matrix shows similar
biological coherence despite the supervised fine-tuning failure,
with distinct immune, signaling, and transcriptional clusters
visible (Supplementary Fig.~\ref{fig:stringgo_bp_coinrich}).
This confirms that the zero-shot embedding space organizes BP
annotations correctly even though the supervised head does not
exploit this organization effectively under the current
fine-tuning procedure.

Cross-ontology sensitivity analysis further shows that GFM
Trained embeddings place proteins sharing the same dominant GO
term in closer neighborhoods than any baseline, consistently
across all three ontologies and all values of $k$ from 25 to 100
(Supplementary Fig.~\ref{fig:stringgo_crossonto}).

These results establish a consistent pattern across three datasets
with different node feature modalities: expression-derived features
in SagePPI, confidence-weighted interaction features in
ogbn-proteins, and amino acid composition in StringGO. In all three
cases the pretrained backbone transfers to the target graph without
seeing it during training. What none of these benchmarks tests
is whether the learned representations generalize when the label
categories themselves are entirely unseen, with no overlap between
the classes observed during training and those evaluated at test
time.

\subsection{Pretrained structural embeddings outperform supervised baselines on entirely unseen protein fold classes in Fold-PPI}

\begin{figure}[pth]
\centering
\includegraphics[width=\linewidth, page = 5]{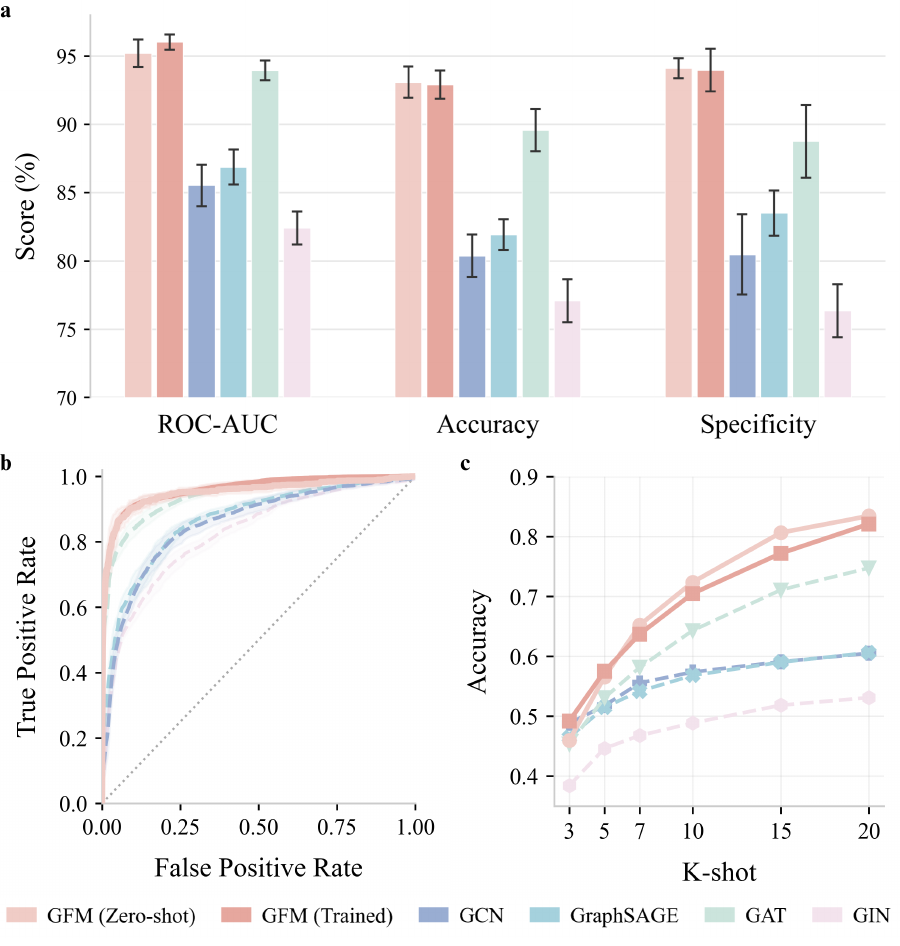}
\caption{\textbf{The pretrained GFM generalizes to unseen protein
structural fold classes across 144 tissue-specific interaction
networks.}
\textbf{a}, Classification performance at $K = 20$ labeled
examples per fold class. Mean ROC-AUC, accuracy, and specificity
are shown for all six models, evaluated on test fold classes that
were entirely absent during training. Error bars denote one
standard deviation across five evaluation seeds.
\textbf{b}, Macro-averaged ROC curves across all test fold
classes. Solid lines denote GFM variants; dashed lines denote
supervised baselines. The dotted diagonal denotes random
performance.
\textbf{c}, Few-shot generalization curves on unseen fold classes.
Mean accuracy is shown as a function of labeled examples per
fold class ($K \in \{3, 5, 7, 10, 15, 20\}$), evaluated on
test fold classes using a logistic probe trained on $K$ support
examples per class. Solid lines denote GFM variants; dashed lines
denote supervised baselines.}
\label{fig:foldppi}
\end{figure}

The previous three benchmarks test generalization across graphs and feature modalities while maintaining at least partial overlap
between the functional categories present during training and those evaluated at test time. Fold-PPI removes this overlap
entirely. It is a graph meta-learning benchmark introduced by
Huang and Zitnik~\cite{huang2020gmeta} comprising 144
tissue-specific protein--protein interaction networks with protein
structural fold class labels drawn from the SCOP
database~\cite{murzin1995scop}. The benchmark uses a strictly disjoint class split in which the fold classes available during training are
entirely absent from the test set, leaving no overlap between the structural categories the model observes and those on which it is
evaluated. This design makes Fold-PPI a direct test of cross-class structural generalization: a model that learns fold-class-specific patterns during training cannot use them to classify test proteins, because test fold classes have no training analog.

The pretrained GFM in zero-shot mode achieved a mean accuracy of 83.4\% and a mean ROC-AUC of 95.5\% on the held-out test
fold classes at $K = 20$ labeled examples per class (Fig.~\ref{fig:foldppi}a,b). GFM Trained reached 82.1\% accuracy
and 95.8\% ROC-AUC under the same protocol. Both values are comparable and substantially exceed the strongest supervised
baseline, GAT~\cite{velickovic2018gat}, which reached 74.8\%
accuracy and 94.1\% ROC-AUC. GCN~\cite{kipf2017gcn} and
GraphSAGE~\cite{hamilton2017inductive} reached 60.5\% and
60.6\% accuracy respectively, and GIN~\cite{xu2019gin} reached
53.1\%. The ROC-AUC gaps between GFM variants and the weaker baselines
are particularly large, exceeding 9 percentage points over GCN and GraphSAGE and 13 percentage points over GIN. Macro-averaged
ROC curves confirm that both GFM variants maintain separation from all baselines across the full threshold range, with the
advantage most pronounced at low false positive rates (Fig.~\ref{fig:foldppi}b).

The few-shot generalization curve reveals how this advantage
builds with label availability (Fig.~\ref{fig:foldppi}c). At
$K = 3$, all models cluster within a narrow accuracy band of
approximately 45--50\%, reflecting the difficulty of classifying
unseen fold classes from very few examples. A clear separation
emerges by $K = 5$, where GFM zero-shot reaches 56.6\% and
GFM Trained reaches 57.5\%, while GAT reaches 53.1\% and GCN
reaches 51.8\%. This gap widens monotonically as $K$ increases.
At $K = 20$, both GFM variants exceed GAT by more than 7
percentage points and exceed GCN, GraphSAGE, and GIN by more
than 20 percentage points. The trajectories of GCN and GraphSAGE
flatten substantially above $K = 10$, indicating that additional
labeled examples provide diminishing returns for models whose
embeddings do not organize the test fold class structure
effectively. The GFM few-shot curves continue to rise steeply
through $K = 20$, consistent with the interpretation that
pretrained topology-based representations provide a strong basis
for fold class discrimination even under the disjoint label
setting.

A notable feature of the Fold-PPI result is that GFM zero-shot
accuracy is marginally higher than GFM Trained accuracy at
$K = 20$ (83.4\% versus 82.1\%), which reverses the relationship
observed in SagePPI, ogbn-proteins, and StringGO. This inversion
is consistent with the disjoint class structure of the benchmark.
Supervised fine-tuning on Fold-PPI trains the backbone to encode
the 19 training fold classes, which introduces class-specific
geometric structure that does not directly transfer to the 5 test
fold classes. GFM zero-shot embeddings, derived from structural
topology alone, are not biased toward any specific fold class and
generalize more uniformly across the class boundary. This
observation reinforces the argument that topology-based
pretraining produces representations that are broadly
transferable precisely because they do not encode dataset-specific
label structure.

t-SNE visualization of GFM Trained embeddings shows that the
29 training fold classes are organized into compact, well-separated
clusters in the embedding space, with one cluster per SCOP fold
class label (Supplementary Fig.~\ref{fig:foldppi_tsne}). Each
cluster corresponds to a distinct structural class as defined
by SCOP secondary structure composition and topology. The
clustering emerges from supervised fine-tuning of the backbone
on training fold classes and demonstrates that the pretrained
model can learn fold-class geometry when label information is
available. Baseline models show substantially less compact and
separated cluster structure under the same visualization, with
GCN and GraphSAGE embeddings producing diffuse distributions
that do not resolve individual fold classes clearly.

GFM Trained embeddings also capture a tissue-prevalence gradient
that is not present in baseline embeddings
(Supplementary Fig.~\ref{fig:foldppi_central}). The
centralization strength, defined as the negative Spearman
correlation between fold class tissue prevalence and mean
centroid distance, reaches 0.296 for GFM Trained, indicating
that proteins belonging to fold classes expressed across more
tissue contexts are placed closer to their class centroid in
embedding space. GAT achieves the highest baseline centralization
strength at 0.126, which is less than half the GFM value.
GCN achieves 0.027, GraphSAGE 0.075, and GIN 0.096. This
gradient reflects the known relationship between structural fold
prevalence and network centrality in tissue-specific PPI graphs:
proteins with broadly expressed structural architectures tend to
occupy topologically central positions across tissue contexts~\cite{jeong2001lethality},
and GFM Trained embeddings encode this positional signal more
faithfully than embeddings derived from supervised training alone~\cite{jeong2001lethality}.

The Fold-PPI benchmark imposes a uniquely demanding generalization
requirement: the fold classes evaluated at test time share no
overlap with those observed during training, and the interaction
networks span 144 distinct tissue contexts that the model never
encounters during pretraining. Under these conditions, GFM
zero-shot outperforms the best supervised baseline at every value
of $K$ without any fold-class-specific training, and the gap over
weaker baselines exceeds 20 percentage points at $K = 20$. The
tissue prevalence gradient and t-SNE cluster structure confirm
that GFM embeddings encode biologically coherent fold-level
organization from the interaction topology, consistent with the
findings on SagePPI, ogbn-proteins, and StringGO. Taken together,
these results demonstrate that structural pretraining produces
representations that generalize to entirely new protein fold
categories defined by SCOP structural classification, without
any fold-class-specific adaptation.

\section{Discussion}

The central finding of this work is that a model pretrained
exclusively on non-biological graphs transfers to protein
interaction networks with no task-specific retraining. The
pretraining set comprises academic citation networks (Cora,
CiteSeer, DBLP, ogbn-arxiv), scientific co-authorship networks
(CoauthorCS, CoauthorPhysics), a Wikipedia-derived computer
science hyperlink network (WikiCS), and e-commerce product
co-purchase networks (AmazonComputers, AmazonPhoto). None of
these graphs contain proteins, biological interactions, or
functional annotations. Despite this, the frozen pretrained
backbone outperforms supervised GNN baselines trained directly
on the target biological graphs across all four evaluation
benchmarks. This outcome is not a marginal effect. On StringGO,
the gap over GCN exceeds 30\% in mean ROC-AUC
(Figs.~\ref{fig:stringgo}a,b). On Fold-PPI, the gap over GIN exceeds 13\% at
$K = 20$ (Fig.~\ref{fig:foldppi}a). On SagePPI, the pretrained model matches what
supervised baselines need 20 labeled examples per process to
approach, using only a single labeled example. These results
indicate that the structural properties used during pretraining
encode organizational principles that are common to graphs
regardless of their domain, and that protein interaction networks
share enough of this organizational structure with citation and
co-purchase networks for the learned representations to transfer
directly.

The mechanism behind this transfer lies in what the structural
prompts encode. Degree, clustering coefficient, $k$-core number,
ego-network statistics, PageRank, and community membership are
defined purely from graph topology. They carry no reference to
what nodes represent or what edges mean. A high-degree node in
a citation network and a high-degree node in a protein
interaction network occupy structurally analogous positions in
their respective graphs, even though one represents a scientific
paper and the other represents a hub protein. The contrastive
pretraining objective aligns text-derived structural descriptions
with graph-derived neighborhood embeddings across nine
structurally diverse graphs simultaneously. This forces the
backbone to learn representations that are sensitive to
topological role rather than to any single graph's semantics.
When applied to biological networks, this topology-sensitive
representation space preserves properties that have biological
meaning because biological function is itself organized
topologically. Hub proteins are functionally central. Community
structure in protein interaction networks reflects shared pathway
membership~\cite{girvan2002community}. Diffusion signatures capture regulatory reach. The
model does not know any of this explicitly. It recovers these
patterns because the structural signals it learned from citation
and co-purchase networks are the same signals that carry
functional meaning in protein networks (Figs.~\ref{fig:sageppi}--\ref{fig:foldppi}).

The GO hierarchy depth result from ogbn-proteins provides the
clearest mechanistic evidence for this argument
(Fig.~\ref{fig:ogbnproteins}e). GFM Trained
performance increases monotonically from 75.4\% on shallow GO
terms to 76.7\% at medium depth and 80.5\% at
deep terms, while GIN shows no consistent
trend across the same stratification. Shallow GO terms annotate
many proteins with broad functions and can be predicted from
local neighborhood patterns alone. Deep GO terms annotate few
proteins with highly specific functions and require integrating
information across long network paths and community boundaries
to identify. The structural prompts encode exactly these long-range
signals through diffusion statistics and multi-scale community
indicators. A model trained on a single graph with a fixed label
set does not acquire this capacity from the training data because
the label distribution during training does not reward long-range
integration in the same way. The depth stratification result
therefore indicates not just that GFM performs better on harder
tasks, but that it performs better for the right structural reason.

The Fold-PPI result reveals a further property of topology-based pretraining that has practical implications. GFM zero-shot marginally outperforms GFM Trained at $K = 20$ on Fold-PPI, reversing the relationship observed on every other benchmark (Fig.~\ref{fig:foldppi}a,c). This inversion is mechanistically consistent with the disjoint
class structure of the benchmark. Supervised fine-tuning on the
19 training fold classes introduces geometric bias toward those
specific structural categories. When the evaluation shifts to
5 entirely unseen fold classes, this bias works against
generalization. GFM zero-shot embeddings carry no class-specific
bias and generalize more uniformly across the class boundary.
This observation has a practical implication for deployment: in
settings where the label vocabulary at test time may differ from
the label vocabulary at training time, topology-based pretrained
representations provide a safer initialization than representations
obtained by supervised fine-tuning on a fixed label set. This
situation arises regularly in biology and medicine, where new
disease categories, newly annotated gene functions, and newly
characterized cell types are continuously introduced after
models are trained.

The multifunctionality gradient provides an independent
validation of the representation space that does not depend on
classification metrics. Across both SagePPI and ogbn-proteins,
GFM zero-shot places highly annotated proteins in denser
embedding neighborhoods (Spearman $r = +0.263$ and $r = +0.105$
respectively; Figs.~\ref{fig:sageppi}b, \ref{fig:ogbnproteins}c), while GIN produces negative correlations on both
graphs ($r = -0.290$ and $r = -0.020$). This pattern is
consistent with the known relationship between multifunctionality
and network centrality in protein interaction networks: proteins
that participate in many biological processes tend to occupy
high-degree, high-betweenness positions that are well connected
across functional modules~\cite{jeong2001lethality,yu2007bottlenecks}. GFM zero-shot captures this
centrality signal directly from topology without any supervision
from functional labels. Supervised baselines do not, because
their training objective optimizes label prediction rather than
structural position. The consistency of the positive GFM
correlation across two structurally different graphs with
different annotation systems supports the interpretation that
this property reflects something real about the learned
representation space rather than an artifact of a particular
dataset.

Co-enrichment analysis across all four benchmarks confirms that
the GFM embedding space recovers known biological relationships
without being told about them. In SagePPI, transcriptional
processes cluster together and immune processes form a separate
coherent block (Fig.~\ref{fig:sageppi}c). In ogbn-proteins, an immune module and a cytokine
module show strong mutual enrichment consistent with their known
coupling in inflammatory signaling. In StringGO, the
cellular component ontology resolves into three spatially
distinct modules corresponding to mitochondria, the nuclear and
cytoplasmic compartment, and the plasma membrane (Fig.~\ref{fig:stringgo}f). These are the
major spatial compartments of eukaryotic cells. They emerge from
protein interaction network topology alone, without any
annotation of subcellular localization provided to the model.
In the molecular function ontology, transcription factor activity
terms cluster according to their shared mechanistic role in
RNA polymerase II-driven gene expression(Fig.~\ref{fig:stringgo}e). None of these
biological relationships are encoded in the structural prompts
or the pretraining objective. They emerge because proteins that
share biological function tend to interact preferentially with
each other, and the learned representations preserve this
interaction-based proximity faithfully.

The BP ontology in StringGO reveals an instructive boundary
condition. GFM zero-shot achieves 80.5\% mean
ROC-AUC on BP, exceeding all supervised baselines
(Supplementary Fig.~\ref{fig:stringgo_bp}), and the
co-enrichment analysis confirms that the zero-shot embedding
space organizes BP annotations into biologically coherent
clusters (Supplementary Fig.~\ref{fig:stringgo_bp_coinrich}). The GFM
Trained model, however, does not improve over zero-shot on this
ontology. The BP label space is substantially larger and more
heterogeneous than MF or CC, and the validation signal from its
imbalanced annotation density creates more instability during
the two-stage supervised fine-tuning procedure used here. This
observation points toward a productive direction: the pretrained
representations encode the necessary functional organization,
and approaches such as label-balanced sampling, ontology-aware
curriculum scheduling, or separate adaptation heads for large
heterogeneous label spaces are well positioned to recover this
gap. The result thus identifies a specific regime where improved
adaptation procedures will be most valuable, rather than a
fundamental limitation of the representations themselves.

The structural prompt framework performs best when graph topology
is meaningfully organized, as it is in interaction networks,
citation graphs, and co-authorship graphs. In settings where
edges are sparse or noisy, the community detection and diffusion
statistics that underpin the prompts will be less reliable, and
the adapter tuning step may benefit from noise-robust community
detection methods. The framework also requires access to full
graph topology at adaptation time, which points toward
future work on inductive structural prompting for streaming or
privacy-constrained graph settings where the complete adjacency
structure is not available at inference time.

The practical significance of this work extends beyond the specific
benchmarks evaluated here. In biology and medicine, labeled data
is typically scarce and expensive to obtain. A model that can
be pretrained once on publicly available non-biological graphs
and then transferred to biological networks with minimal
adaptation reduces the labeled data requirement substantially.
The few-shot results across all four benchmarks demonstrate
this concretely: the pretrained backbone provides useful
representations from as few as one labeled example per class.
This is particularly relevant for emerging biological data types
where functional annotations are incomplete, such as newly
sequenced organisms, newly characterized cell types in
single-cell atlases~\cite{regev2017humancellatlas}, and newly constructed disease-specific
interaction networks where curated labels are not yet available.
The structural prompt framework is not specific to protein
interaction networks. The same prompts can be computed for any
graph with meaningful topology, including gene regulatory
networks, cell-cell communication graphs, drug-target interaction
networks~\cite{stokes2020antibiotic}, and clinical knowledge graphs. Whether the
representations learned from citation and co-purchase networks
transfer to these other biological and clinical graph types
remains an open question that this work positions for direct
experimental investigation.

\subsection*{Outlook}

We have introduced a graph foundation model that learns
transferable structural representations from nine heterogeneous
non-biological graphs and applies them directly to protein
interaction networks without retraining. The model encodes
each node's structural role through topology-derived prompts
covering local connectivity, global community membership,
and diffusion-based position, and aligns these descriptions
with graph-derived neighborhood embeddings through contrastive
pretraining. Across four biological network benchmarks spanning
human protein interaction networks, a large-scale species-stratified
STRING-derived benchmark, a multi-ontology STRING network with
amino acid composition features, and a collection of 144
tissue-specific interaction networks with disjoint structural
fold class labels, the frozen pretrained backbone consistently
matches or exceeds supervised GNN baselines trained directly
on the target graphs. In zero-shot mode, it outperforms the
best supervised baseline on all four benchmarks. In few-shot
mode, it achieves with a single labeled example per class what
supervised models require up to 20 labeled examples to approach.
The learned embedding space recovers known biological
relationships including transcriptional modules, immune and
cytokine signaling clusters, mitochondrial compartment
organization, and cellular localization boundaries without
any supervision from functional annotations. Performance
improves with the specificity of GO annotations, indicating
that the topology-based representations encode the long-range
network signals that are most informative for fine-grained
functional discrimination. On a benchmark with entirely unseen
protein fold classes at test time, the zero-shot model
outperforms supervised fine-tuning, demonstrating that
topology-based pretraining produces representations that
generalize across label boundaries rather than within them.
These results establish that structural graph properties are
a viable foundation for cross-domain graph representation
learning, and that a single model pretrained on publicly
available non-biological graphs can serve as a practical
starting point for biological network analysis across
diverse graph types, annotation systems, and evaluation regimes
where labeled data is limited.

\section{Methods}

\subsection{Overview}

The Graph Foundation Model (GFM) is designed to learn node
representations that transfer across graphs with incompatible
node feature spaces. The central challenge is that standard
graph neural networks encode node features directly, which
means the learned parameters are specific to the feature
schema of the training graph and cannot be reused on a graph
with a different feature type or dimensionality. The GFM
addresses this by replacing raw node features with a
feature-agnostic structural description of each node's
position and role in the graph. This description is derived
entirely from topology and is expressible as a natural language
string that can be encoded by a language model into a
fixed-dimensional vector, regardless of the original feature
space. The model is pretrained once by aligning these
topology-derived text embeddings with graph-structure embeddings
produced by a message-passing backbone across nine heterogeneous
graphs simultaneously. After pretraining, the backbone is frozen
and reused on unseen graphs by fitting a lightweight per-graph
adapter that maps the new graph's node features into the
pretrained representation space. This adapter is tuned without
labels using a contrastive objective on the target graph's
topology. The resulting embeddings can then be used directly
with a linear probe for zero-shot evaluation or fine-tuned with
a task-specific head when labeled data is available.

\subsection{Structural prompt construction}

For each node $i$ in a graph $G = (V, E)$, a structured
natural-language profile is constructed from two categories
of topological descriptors: local context and global context.

Local context descriptors capture the immediate structural
environment of the node. These include the node degree $d_i$,
the local clustering coefficient $c_i$, the $k$-core
number~\cite{batagelj2003core}, first- and second-order
ego-network statistics (vertex count, edge count, and density
at radius 1 and 2), and the PageRank score $p_i$ computed with
damping factor 0.85 over 40 power iterations~\cite{page1999pagerank}.

Global context descriptors capture the node's position within
the large-scale community structure of the graph. Two
independent community detection algorithms are applied. Label
Propagation~\cite{raghavan2007near} is run for 20 iterations
to assign each node to a community, and the community identity,
size, and internal edge density are recorded. SCoDA, a streaming
community detection algorithm~\cite{hollocou2017scoda}, is applied independently
and its community identity, size, and density are recorded
separately. Using two algorithms reduces dependence on any
single community detection method and provides complementary
views of mesoscale structure.

These descriptors are assembled into a natural-language node
profile of the form:

\begin{quote}
\small
\texttt{Node profile: local(deg=12, cc=0.167, core=4,
ego1V=13, ego1E=22, ego1D=0.046, \\
ego2V=214, ego2E=4103, ego2D=0.181, pr=0.00084); global(lp\_comm=5, lp\_size=312, \\
lp\_dens=0.021; scoda\_comm=8, scoda\_size=189,
scoda\_dens=0.018); graph(N=2708, E=5429, avgd=4.01,
trans=0.241, q25=2, q50=3, q75=5, spec\_gap=1.23).}
\end{quote}

Each profile is tokenized using the all-MiniLM-L6-v2
Sentence Transformer~\cite{reimers2019sentencebert}, producing
a 384-dimensional embedding $z_i^{\text{text}}$ for each node.
The full computational definitions of each descriptor are
provided in Supplementary Methods.

\subsection{Graph Feature Embedding}

For a graph with $N$ nodes and raw node feature matrix
$X \in \mathbb{R}^{N \times d}$, a per-graph adapter
$A: \mathbb{R}^{N \times (d + 384)} \to \mathbb{R}^{N \times 1024}$
is applied to the concatenation of raw features and structural
tokens:

\begin{equation}
\tilde{X} = A\bigl([X \,\|\, Z^{\text{text}}]\bigr)
\end{equation}

where $Z^{\text{text}} \in \mathbb{R}^{N \times 384}$ is the
matrix of MiniLM token embeddings and $\|$ denotes row-wise
concatenation. The adapter is a single linear layer with output
dimension 1024. A separate adapter is instantiated for each
graph during pretraining. For held-out graphs, a new adapter
is initialized from the mean of the pretrained adapter weights
and tuned without labels on the target graph's topology.

The adapted features $\tilde{X}$ are processed by a two-layer
GraphSAGE backbone~\cite{hamilton2017inductive} with hidden
dimension 512 and output dimension 256, using dropout rate 0.6
at each layer:

\begin{equation}
g = \text{Normalize}\bigl(\text{GraphSAGE}(\tilde{X}, E)\bigr)
\in \mathbb{R}^{N \times 256}
\end{equation}

A text projection head maps the same adapted features to a
matching space:

\begin{equation}
z = \text{Normalize}\bigl(W_{\text{proj}} \tilde{X}\bigr)
\in \mathbb{R}^{N \times 256}
\end{equation}

where $W_{\text{proj}} \in \mathbb{R}^{1024 \times 256}$. The
final node embedding is a weighted concatenation of the graph
stream and the text stream:

\begin{equation}
e_i = [\alpha \cdot g_i \,\|\, (1 - \alpha) \cdot z_i]
\in \mathbb{R}^{512}
\end{equation}

with $\alpha = 0.7$.

\subsection{Contrastive pretraining objective}

The model is pretrained by aligning the graph structure
embedding $g_i$ with the text stream embedding $z_j$ of
structurally related nodes. Positive pairs $(i, j)$ are
identified using Personalized PageRank (PPR) with restart
probability 0.15, run for 100 iterations per anchor node.
The top-96 highest-PPR nodes relative to each anchor are
treated as positives~\cite{gasteiger2019predict}. This
selection captures multi-hop structural proximity without
relying on direct adjacency.

The contrastive loss is an InfoNCE
objective~\cite{oord2018representation,chen2020simclr} with temperature
$\tau = 0.1$:

\begin{equation}
\mathcal{L}_{\text{NCE}} = -\frac{1}{2} \Bigl(
\log \frac{\exp(g_i \cdot z_j / \tau)}
{\sum_{k} \exp(g_i \cdot z_k / \tau)}
+
\log \frac{\exp(z_j \cdot g_i / \tau)}
{\sum_{k} \exp(z_j \cdot g_k / \tau)}
\Bigr)
\end{equation}

averaged symmetrically over both directions. For graphs with
more than 20,000 nodes, a sampled variant with 1024 in-batch
negatives is used to remain memory-efficient. For smaller
graphs the full node bank is used as the negative set,
identical to the original InfoNCE formulation.

A Laplacian smoothing regularizer penalizes large differences
between embeddings of adjacent nodes:

\begin{equation}
\mathcal{L}_{\text{smooth}} = \lambda \cdot
\frac{1}{|E|} \sum_{(u,v) \in E} \|g_u - g_v\|^2
\end{equation}

with $\lambda = 5 \times 10^{-3}$. The total training loss is
$\mathcal{L} = \mathcal{L}_{\text{NCE}} + \mathcal{L}_{\text{smooth}}$.

\subsection{Pretraining protocol}

The model is pretrained on nine heterogeneous graphs spanning
academic citation networks (Cora~\cite{yang2016revisiting},
CiteSeer~\cite{yang2016revisiting},
DBLP~\cite{yang2016revisiting},
ogbn-arxiv~\cite{hu2020ogb}),
scientific co-authorship networks
(CoauthorCS~\cite{shchur2018pitfalls},
CoauthorPhysics~\cite{shchur2018pitfalls}),
e-commerce product co-purchase networks
(AmazonComputers~\cite{shchur2018pitfalls},
AmazonPhoto~\cite{shchur2018pitfalls}),
and a Wikipedia computer science hyperlink network
(WikiCS~\cite{mernyei2020wiki}).
None of these graphs contain proteins, biological interactions,
or functional annotations.

At each training step, one graph is sampled uniformly at
random. A batch of 1024 anchor nodes is drawn from that graph,
PPR positives are computed, and the loss is evaluated on the
anchor-positive pairs. Training proceeds for 250 epochs with
128 steps per epoch. The Adam optimizer~\cite{kingma2015adam}
is used with learning rate $10^{-5}$ and weight decay
$5 \times 10^{-4}$. Mixed-precision training~\cite{micikevicius2018mixed} is applied
throughout. The checkpoint with the lowest training loss is
retained for downstream evaluation.

\subsection{Downstream adaptation}

For each evaluation graph, a new linear adapter is initialized
from the mean of the pretrained adapter weights and tuned
without labels using the same contrastive objective applied to
the target graph's topology. For SagePPI the adapter is tuned
for 2000 steps. For ogbn-proteins, StringGO, and Fold-PPI the
adapter is tuned for 5000 steps. The backbone and text
projection head remain frozen during adapter tuning.

Zero-shot evaluation uses the frozen backbone with the adapted
embeddings and a logistic regression linear probe fitted on the
training split. Supervised fine-tuning proceeds in two stages.
In the first stage, the backbone and text projection are frozen
and only the adapter and task-specific classification head are
trained. In the second stage, all parameters are unfrozen and
trained jointly with a lower learning rate for the backbone
($10^{-5}$) and text projection ($10^{-5}$) than for the
adapter ($10^{-4}$) and head ($10^{-3}$). Both stages use
binary cross-entropy loss and are monitored on the validation
split by mean ROC-AUC. The best validation checkpoint is
retained for test evaluation.

Few-shot evaluation trains a logistic regression probe on
$K$ labeled examples per class drawn from the training split,
with $K \in \{1, 5, 10, 20\}$ for SagePPI and ogbn-proteins,
$K \in \{1, 5, 10, 20\}$ for StringGO, and
$K \in \{3, 5, 7, 10, 15, 20\}$ for Fold-PPI. For each value
of $K$, the probe is evaluated on the full test split.

\subsection{Evaluation datasets}

\subsubsection{SagePPI}

The SagePPI dataset is derived from the protein--protein
interaction network introduced alongside the GraphSAGE
framework~\cite{hamilton2017inductive}. It comprises 24
tissue-specific human PPI graphs with a total of 56,944 nodes
and 818,716 edges, averaging approximately 2,372 proteins per
graph. The standard split allocates 20 graphs to training,
2 to validation, and 2 to testing. Node features are
50-dimensional vectors composed of positional gene sets, motif
gene sets, and immunological signatures. Multilabel
classification targets are 121 binary labels derived from the
MSigDB Gene Ontology biological process
collection~\cite{liberzon2015msigdb}.

\subsubsection{ogbn-proteins}

The ogbn-proteins benchmark is from the Open Graph
Benchmark~\cite{hu2020ogb}. The underlying network is derived
from the STRING database version 11~\cite{szklarczyk2019string}.
It contains 132,534 protein nodes across eight species connected
by 39,561,252 undirected edges representing biologically
meaningful associations including physical interactions and
genetic co-expression. The original dataset contains no node
features. Node representations are constructed by aggregating
the 8-dimensional edge features of each node's incident edges.
Multilabel classification targets are 112 binary GO function
labels. The benchmark uses a species-stratified split in which
test proteins are drawn from species absent during training.
No modifications were made to the standard split.

\subsubsection{StringGO}

The StringGO dataset is constructed from the human interactome
in the STRING database~\cite{szklarczyk2019string}. Nodes
represent approximately 18,000 human proteins. Edges represent
predicted and experimental functional associations, yielding
approximately 8,000,000 interactions. Node features are
20-dimensional amino acid composition vectors. Labels are
derived from the Gene Ontology resource~\cite{go2021} and
evaluated separately across three ontologies: Molecular
Function, Biological Process, and Cellular Component. Each
ontology defines its own train/validation/test split over
proteins.

\subsubsection{Fold-PPI}

The Fold-PPI dataset was introduced with the G-Meta graph
meta-learning framework~\cite{huang2020gmeta}. It comprises
144 tissue-specific human protein--protein interaction networks.
Nodes represent proteins and edges represent physical
interactions localized to a specific tissue context. Node
features are 512-dimensional conjoint triad vectors~\cite{shen2007conjoint} encoding
amino acid composition and local sequence triplet statistics.
Labels correspond to 30 protein structural fold classes from
the SCOP database~\cite{murzin1995scop}. The benchmark uses a
strictly disjoint class split in which 19 fold classes are
available during training and 5 entirely distinct fold classes
are reserved for evaluation. No fold class identity is shared
between the training and test sets.

\subsection{Baseline implementations}

Four supervised GNN baselines are evaluated on each dataset:
Graph Convolutional Networks (GCN)~\cite{kipf2017gcn}, Graph
Isomorphism Networks (GIN)~\cite{xu2019gin}, Graph Attention
Networks (GAT)~\cite{velickovic2018gat}, and
GraphSAGE~\cite{hamilton2017inductive}. All four baselines
use a two-layer architecture with hidden dimension 256. Each
baseline is trained from scratch on the target graph using raw
node features only, without structural prompts or pretrained
weights. Training uses the Adam optimizer with learning rate
$5 \times 10^{-3}$, weight decay $5 \times 10^{-4}$, and
dropout rate 0.5. A maximum of 500 training epochs is applied
with early stopping based on validation mean ROC-AUC using a
patience of 500 epochs. GAT uses 4 attention heads in the
first layer and 1 head in the second. The same hyperparameters
are applied across all four evaluation datasets without
dataset-specific tuning.

\subsection{Evaluation metrics}

The primary evaluation metric is mean ROC-AUC, computed as the
unweighted average of per-label ROC-AUC scores across all
valid labels (labels with at least two represented classes in
the test set). Classification accuracy is reported as a
secondary metric using per-label decision thresholds tuned on
the validation set by maximizing binary F1 score. For Fold-PPI,
accuracy is the primary metric consistent with the original
benchmark protocol. All experiments use random seed 42.

\section*{Data availability}
All datasets analyzed in this study are publicly available from
established repositories. The SagePPI protein--protein interaction
network was obtained from the GraphSAGE framework repository
introduced by Hamilton \textit{et al.}~\cite{hamilton2017inductive}.
The ogbn-proteins benchmark was obtained from the Open Graph
Benchmark~\cite{hu2020ogb}. The StringGO dataset was constructed
from the STRING database version 11~\cite{szklarczyk2019string}
with Gene Ontology annotations from the Gene Ontology
Consortium~\cite{go2021}. The Fold-PPI dataset was obtained from
the G-Meta repository introduced by Huang and
Zitnik~\cite{huang2020gmeta}. All nine pretraining graphs (Cora,
CiteSeer, DBLP, ogbn-arxiv, CoauthorCS, CoauthorPhysics,
AmazonComputers, AmazonPhoto, and WikiCS) are publicly available
through the PyTorch Geometric library~\cite{fey2019pyg} and the
Open Graph Benchmark~\cite{hu2020ogb}.

All processed structural token caches, pretrained model checkpoints,
and intermediate embedding files required to reproduce the analyses
reported in this study are provided within the associated Code Ocean
capsule prepared for peer review. Detailed preprocessing procedures
and model training protocols are described in the Methods. No new
human or animal data were generated for this study.

\section*{Code availability}
All code used to perform the analyses in this study has been
deposited in a Code Ocean capsule and shared with the editors and
reviewers as part of the peer-review process. The capsule contains
the complete implementation of the Graph Foundation Model,
including structural prompt construction, multi-graph pretraining,
per-graph adapter tuning, downstream evaluation pipelines, and
all scripts used to generate the figures reported in this study.

Upon acceptance of the manuscript, the Code Ocean capsule will be
made publicly available with a persistent digital object identifier
(DOI) to enable full reproducibility of the results.

\bibliography{main}

\clearpage
\newpage
\appendix
\setcounter{figure}{0}
\renewcommand{\thefigure}{S\arabic{figure}}
\setcounter{table}{0}
\renewcommand{\thetable}{S\arabic{table}}

\section*{Supplementary Materials}
\begin{figure}[pbh]
\centering
\includegraphics[width=0.78\linewidth, page = 6]{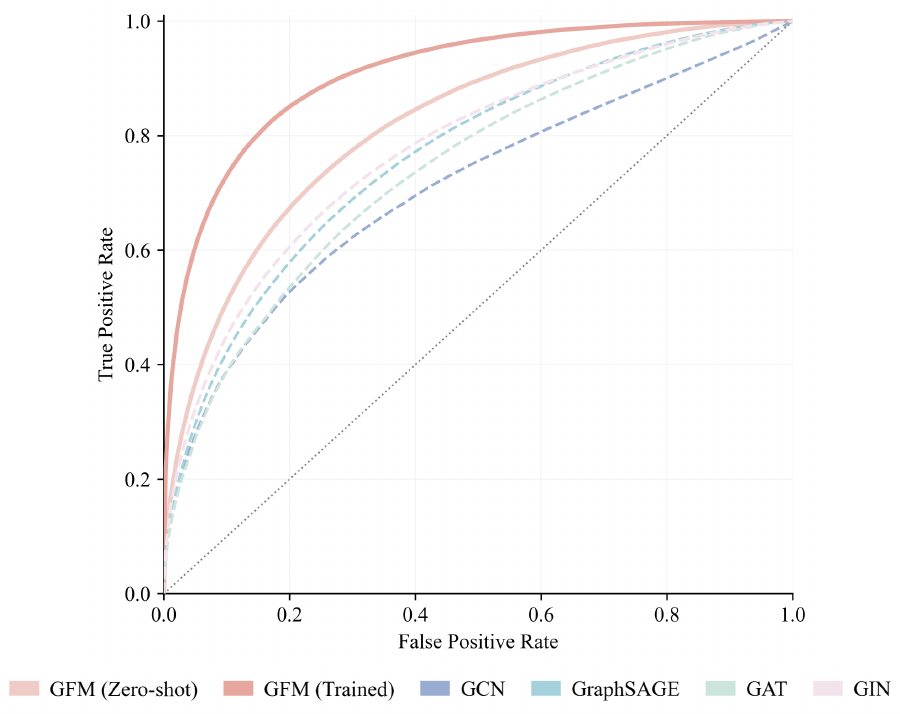}
\caption{\textbf{Full macro-averaged ROC curves for SagePPI.}
Macro-averaged receiver operating characteristic curves across all 121
GO biological process labels for GFM zero-shot, GFM Trained, GCN,
GraphSAGE, GAT, and GIN. Curves are computed by interpolating per-label
ROC curves onto a common false positive rate grid and averaging. The
clear separation between GFM variants (solid lines) and all supervised
baselines (dashed lines) is maintained across the full operating range,
with the largest absolute gap occurring at low false positive rates
where classifier precision requirements are most stringent.}
\label{fig:roc_full}
\end{figure}


\begin{figure}[t]
\centering
\includegraphics[width=\linewidth, page = 7]{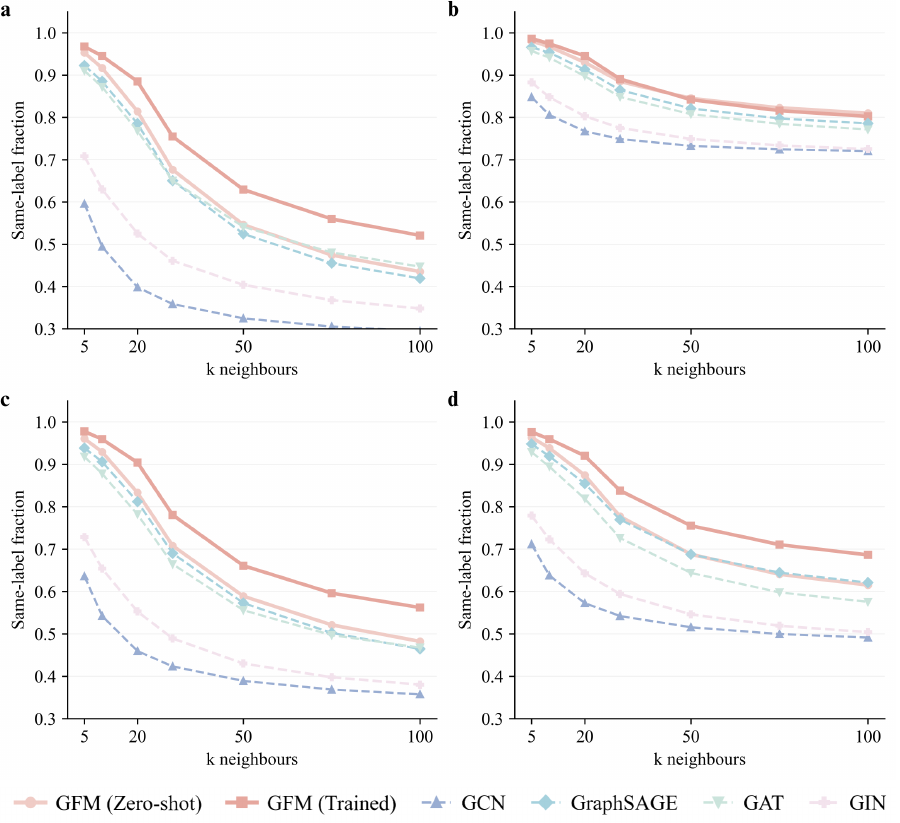}
\caption{\textbf{Same-label neighbor enrichment curves for four GO
biological processes in SagePPI.}
For each model, the same-label fraction is computed as the proportion of
k nearest neighbors (in cosine embedding space) that share the same GO
process annotation as the query protein, averaged over all proteins
annotated with that process. Curves are shown for k $\in$
\{5, 10, 20, 30, 50, 75, 100\}.
\textbf{a}, DNA repair (GO:0006281).
\textbf{b}, Cell cycle regulation (GO:0051726).
\textbf{c}, Nervous system development (GO:0007399).
\textbf{d}, Negative regulation of transcription (GO:0000122).
Solid lines denote GFM variants; dashed lines denote supervised baselines.
GFM zero-shot and GFM Trained consistently maintain higher same-label
fractions at all values of k across all four processes.}
\label{fig:enrichment_s2}
\end{figure}


\begin{figure}[t]
\centering
\includegraphics[width=\linewidth, page = 8]{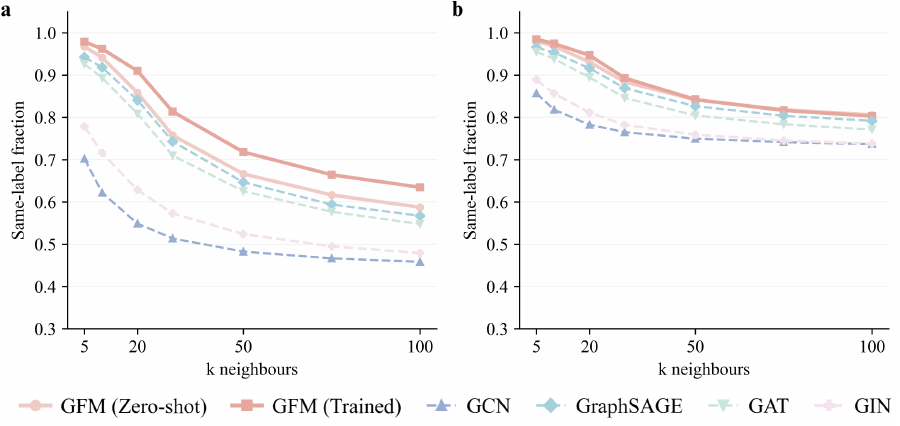}
\caption{\textbf{Same-label neighbor enrichment curves for two additional
GO biological processes in SagePPI.}
\textbf{a}, RNA polymerase II transcription (GO:0006366).
\textbf{b}, Cell proliferation regulation (GO:0042127).
Format and evaluation protocol are identical to Supplementary Fig.~2.
The GFM advantage is maintained across both processes, including cell
proliferation regulation, which represents a functionally distinct
class from the transcriptional and repair processes shown in
Supplementary Fig.~2.}
\label{fig:enrichment_s3}
\end{figure}

\begin{figure}[t]
\centering
\includegraphics[width=0.78\linewidth, page = 9]{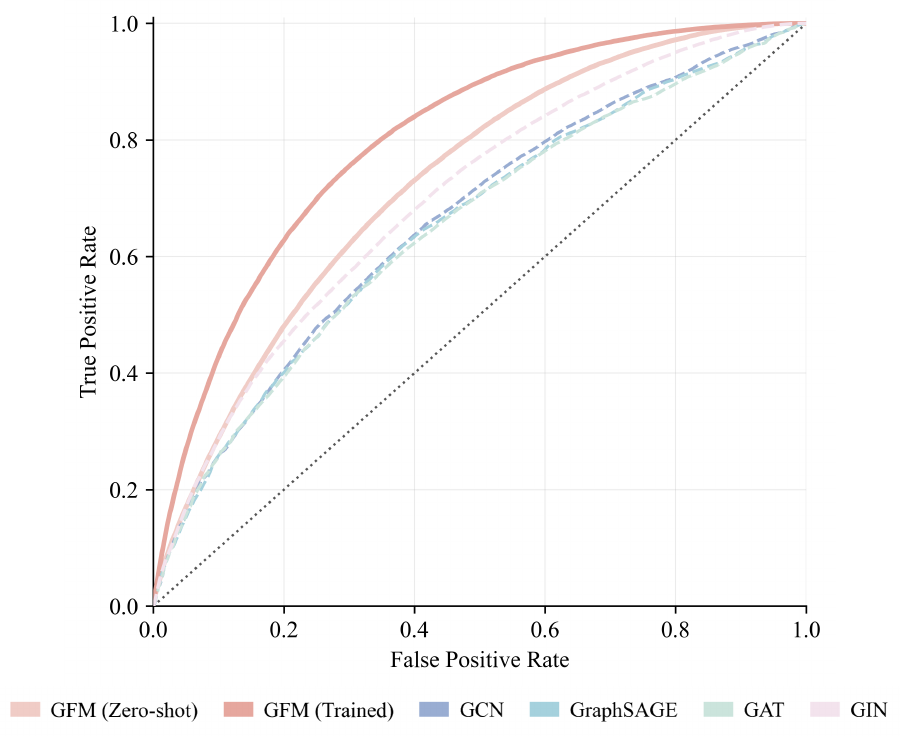}
\caption{\textbf{Full macro-averaged ROC curves for ogbn-proteins.}
Macro-averaged receiver operating characteristic curves across all
GO biological process labels for GFM zero-shot, GFM Trained, GCN,
GraphSAGE, GAT, and GIN. The clear separation between GFM variants
(solid lines) and all supervised baselines (dashed lines) persists
across the full operating range. The gap is widest at low false
positive rates, consistent with results on SagePPI
(Supplementary Fig.~\ref{fig:roc_full}).}
\label{fig:ogbn_roc_full}
\end{figure}


\begin{figure}[t]
\centering
\includegraphics[width=0.85\linewidth, page = 10]{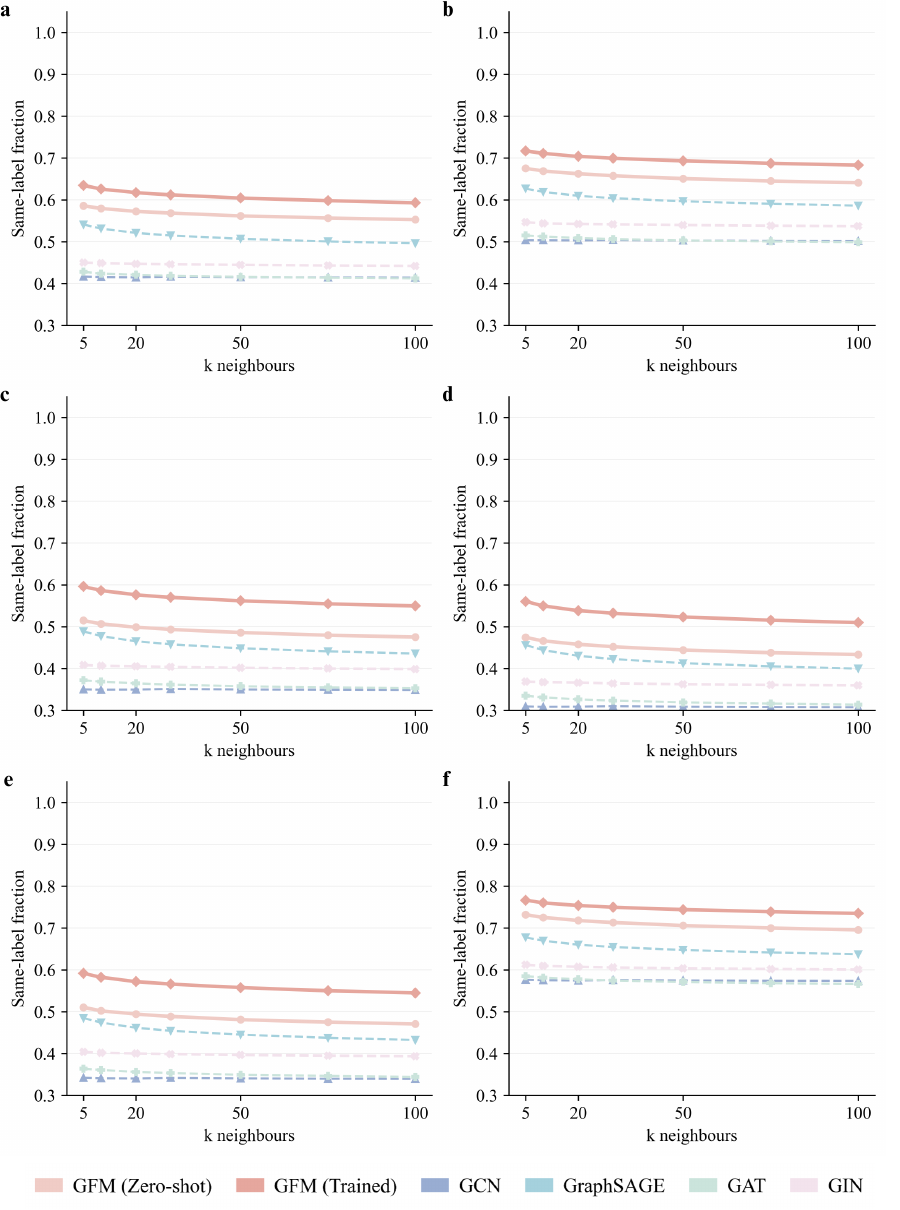}
\caption{\textbf{Same-label neighbor enrichment curves for six GO
annotation categories in ogbn-proteins.}
For each model, the same-label fraction is computed as the proportion
of $k$ nearest neighbors (in cosine embedding space) that share the
same GO annotation as the query protein, averaged over all annotated
proteins. Curves are shown for $k \in \{5, 10, 20, 30, 50, 75, 100\}$.
\textbf{a}, Molecular function.
\textbf{b}, Cellular process.
\textbf{c}, Binding.
\textbf{d}, Organic substance metabolic process (GO:0071704).
\textbf{e}, Cellular metabolic process (GO:0044237).
\textbf{f}, Nitrogen compound metabolic process (GO:0006807).
Solid lines denote GFM variants; dashed lines denote supervised
baselines. GFM zero-shot and GFM Trained consistently maintain
higher same-label fractions at all values of $k$ across all six
annotation categories.}
\label{fig:ogbn_enrichment}
\end{figure}


\begin{figure}[t]
\centering
\includegraphics[width=0.85\linewidth, page = 11]{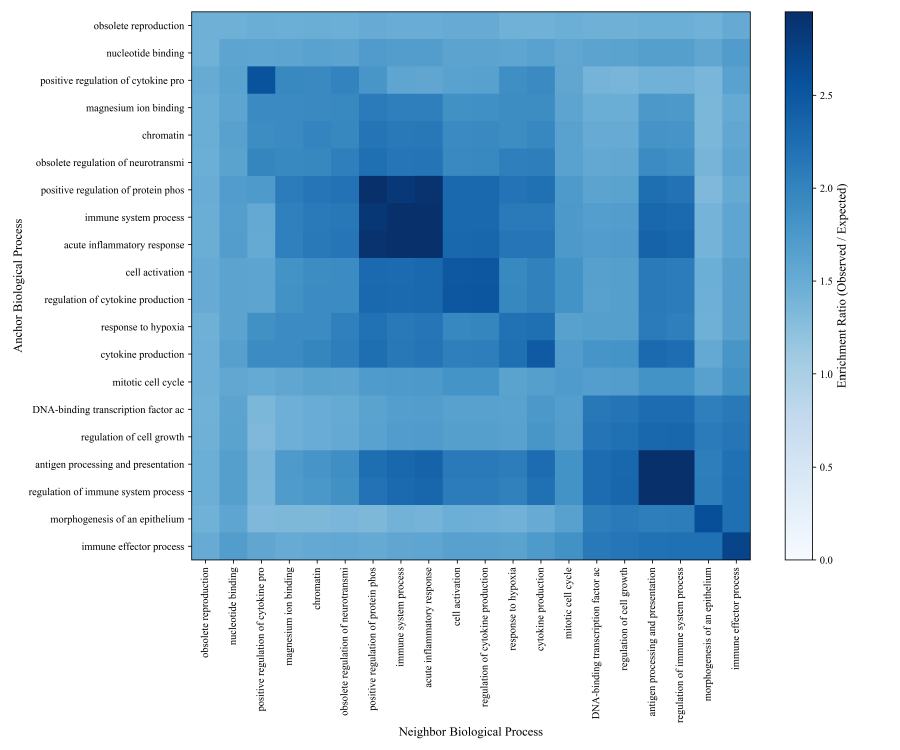}
\caption{\textbf{Cross-functional co-enrichment in GFM Trained
embeddings for ogbn-proteins.}
Each cell reports the enrichment ratio (observed over expected) for
the mean fraction of GO-annotated neighbors that a protein from one
anchor biological process accumulates within a 10-nearest-neighbor
embedding neighborhood annotated by another process. Rows and columns
are ordered by hierarchical clustering of the symmetrized co-enrichment
matrix. Enrichment ratios substantially above 1.0 indicate that two
processes share embedding-space neighborhoods more than expected by
chance. The immune cluster---comprising regulation of immune system
process, antigen processing and presentation, acute inflammatory
response, immune system process, and immune effector process---forms
a prominent high-enrichment block. A separate cytokine cluster---positive
regulation of cytokine production, regulation of cytokine production,
cytokine production, and cell activation---shows strong internal
enrichment and cross-enrichment with the immune block, consistent
with the known coupling between cytokine signaling and acute
inflammatory programs.}
\label{fig:ogbn_coinrich}
\end{figure}


\begin{figure}[t]
\centering
\includegraphics[width=0.75\linewidth, page = 12]{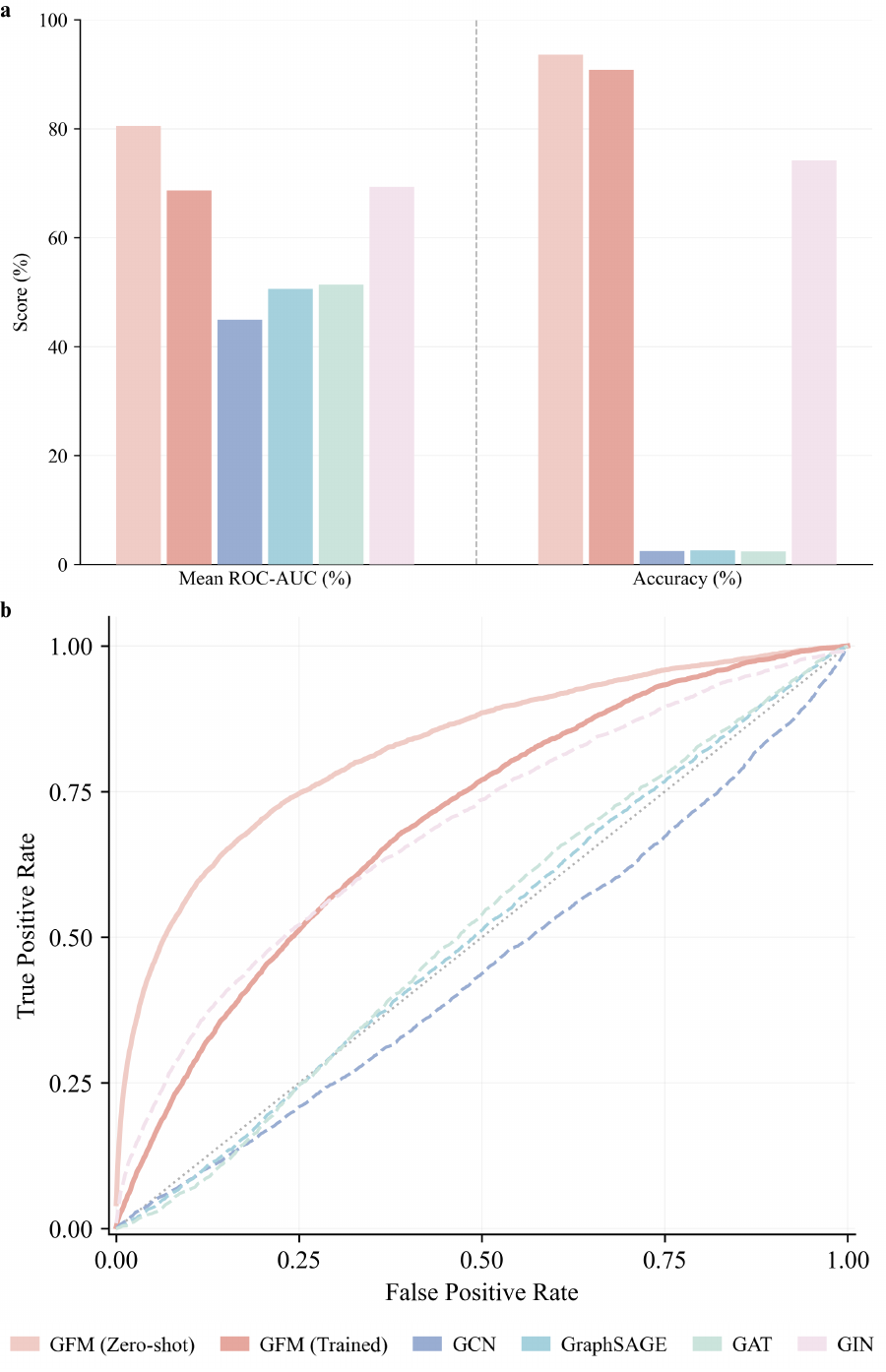}
\caption{\textbf{GFM performance on Biological Process (BP) GO
term prediction in StringGO.}
\textbf{a}, Mean ROC-AUC and accuracy for BP GO term prediction
across all six models. GFM zero-shot achieves the highest mean
ROC-AUC (80.5\%), exceeding all supervised baselines including
GIN (69.3\%). The GFM Trained model does not improve over the
zero-shot baseline on this ontology, in contrast to the MF and
CC results. This behavior is specific to BP and is discussed in
the main text.
\textbf{b}, Macro-averaged ROC curves for BP. Solid lines denote
GFM variants; dashed lines denote supervised baselines. The dotted
diagonal denotes random performance. GFM zero-shot maintains a
clear separation from all baselines across the full threshold
range despite the fine-tuning failure of GFM Trained.}
\label{fig:stringgo_bp}
\end{figure}


\begin{figure}[t]
\centering
\includegraphics[width=\linewidth, page = 13]{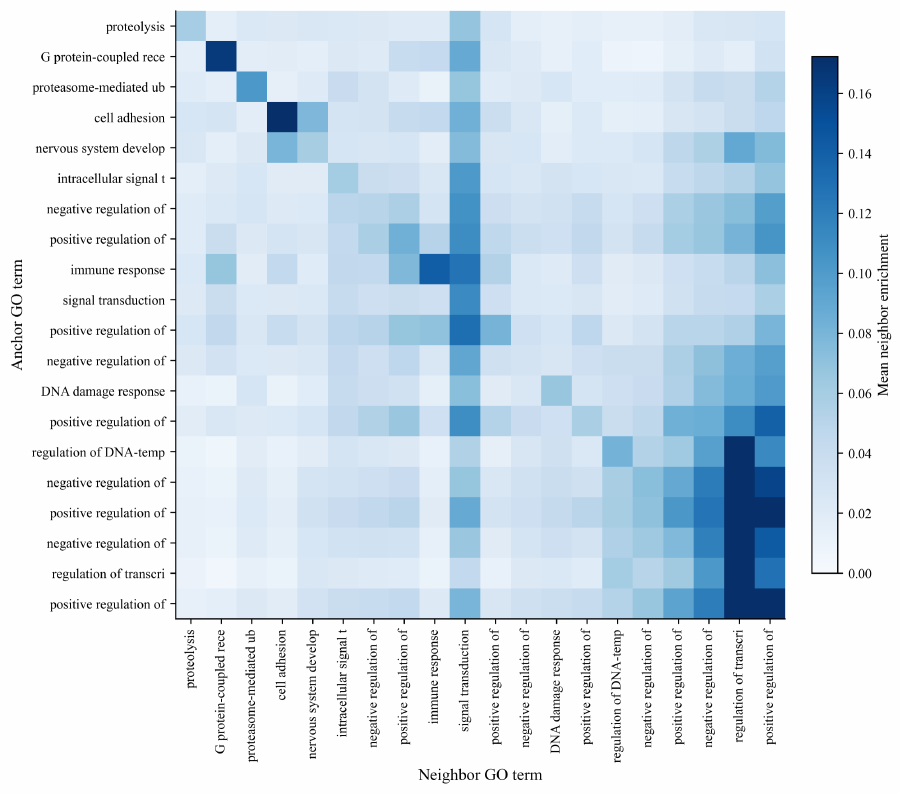}
\caption{\textbf{Noise robustness of GFM and baseline embeddings
in StringGO.}
Mean ROC-AUC is shown as a function of Gaussian noise level
applied to the amino acid composition features ($x_{\text{raw}}$),
for MF (\textbf{a}), CC (\textbf{b}), and BP (\textbf{c}).
Noise is parameterized as a fraction of the per-feature standard
deviation across proteins (0\%, 10\%, 30\%, 50\%).
GFM zero-shot retains a mean ROC-AUC above 0.79 for MF and CC
at 50\% noise, indicating that its embeddings are largely driven
by topology-derived structural features that are not affected by
amino acid feature perturbation. GFM Trained similarly maintains
strong performance under all noise levels for MF and CC. On BP,
both GFM variants show modest decline but remain above all
baselines at every noise level. Baseline performance declines
more steeply because amino acid composition constitutes their
full input.}
\label{fig:stringgo_noise}
\end{figure}


\begin{figure}[t]
\centering
\includegraphics[width=0.75\linewidth, page = 14]{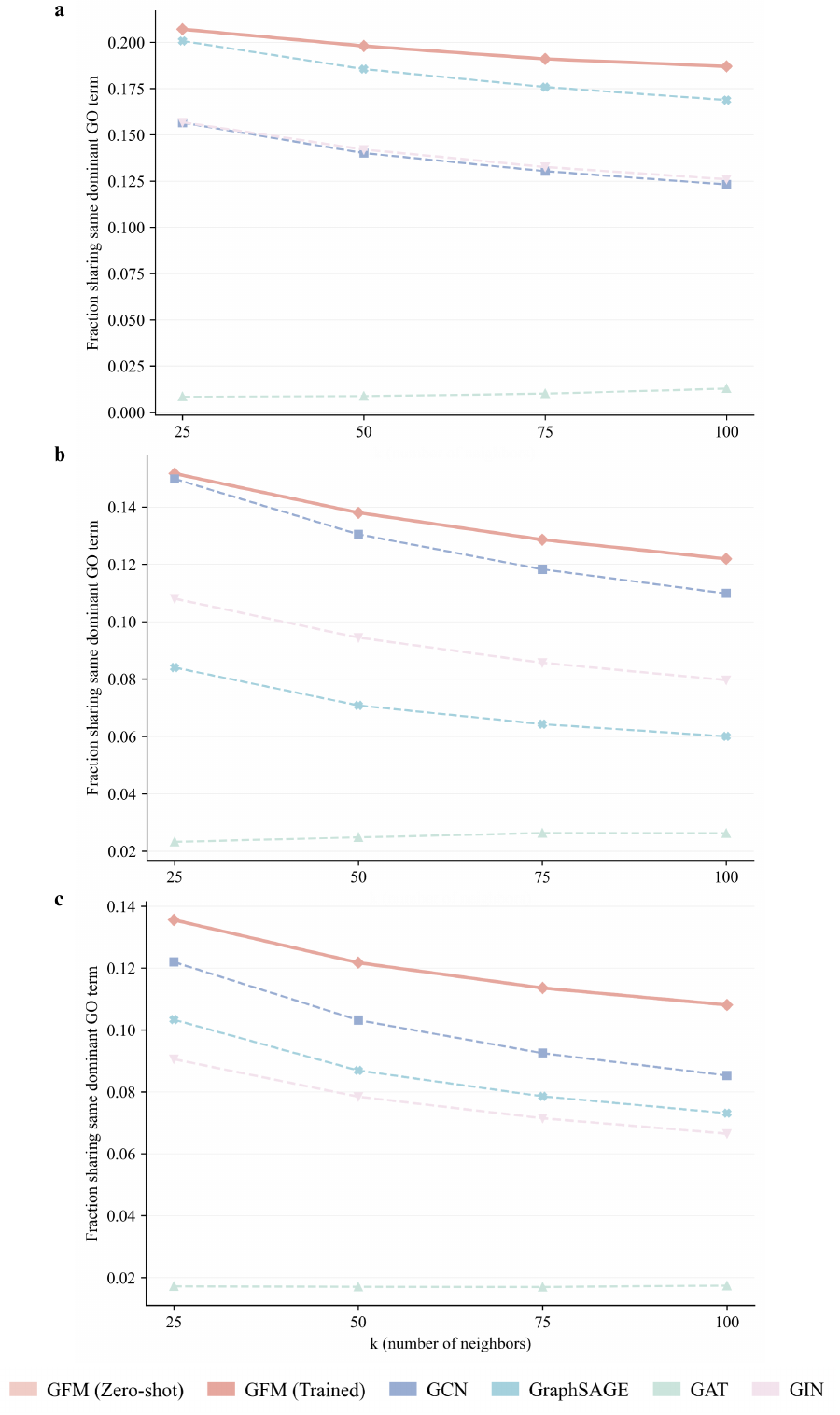}
\caption{\textbf{Cross-functional co-enrichment in GFM Trained
embeddings for BP in StringGO.}
Same protocol as Fig.~\ref{fig:stringgo}e--f. The BP co-enrichment
matrix reveals distinct clusters including an immune response
module, a signal transduction module, and a transcriptional
regulation module. Consistent with the BP performance results,
the co-enrichment structure in the GFM zero-shot embedding space
is well organized even though the supervised head does not
capitalize on it effectively during fine-tuning.}
\label{fig:stringgo_bp_coinrich}
\end{figure}


\begin{figure}[t]
\centering
\includegraphics[width=0.75\linewidth, page = 15]{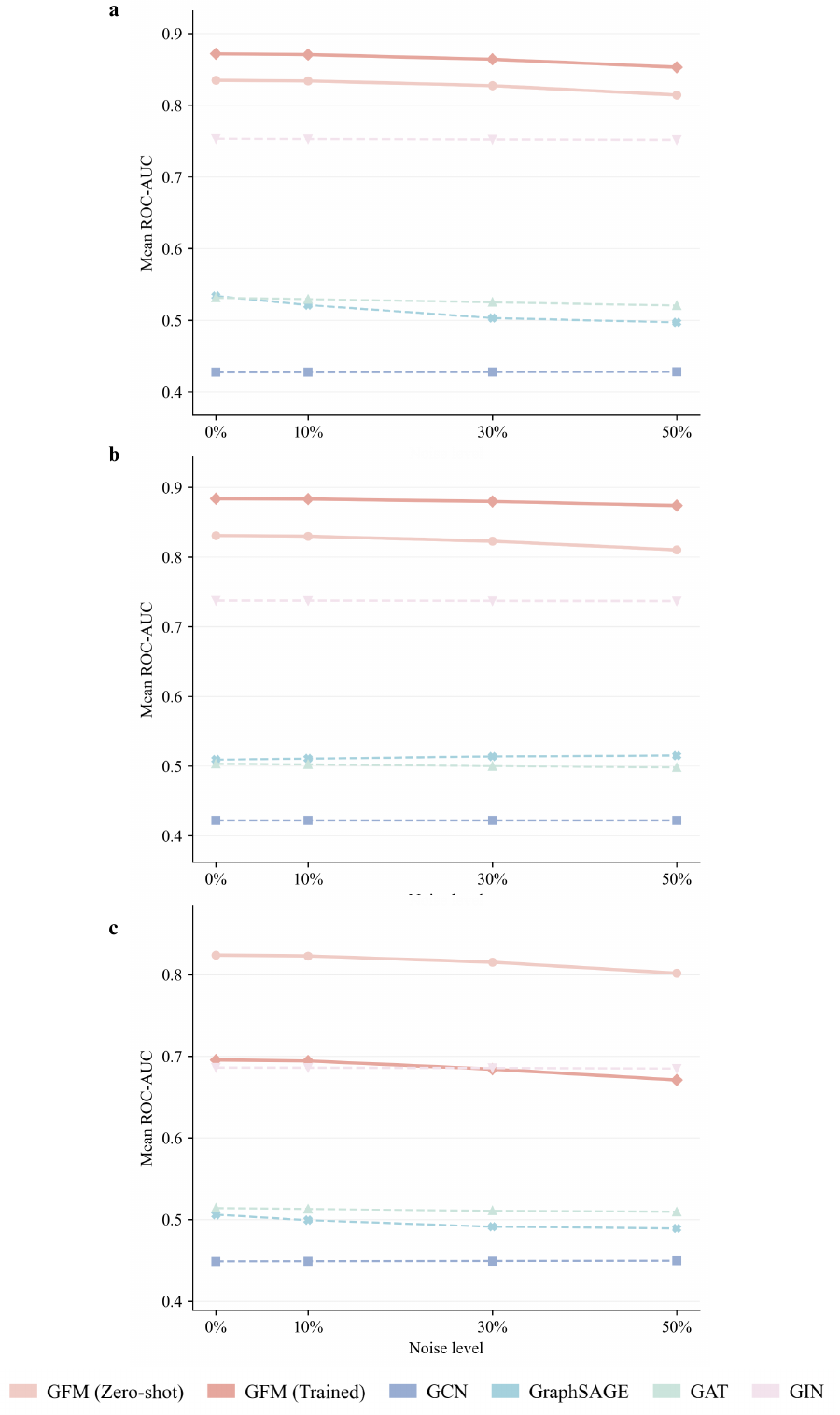}
\caption{\textbf{Cross-ontology neighborhood sensitivity in
StringGO embeddings.}
For each model and ontology, the fraction of $k$ nearest neighbors
sharing the same dominant GO term is shown for
$k \in \{25, 50, 75, 100\}$ across MF (\textbf{a}), BP (\textbf{b}),
and CC (\textbf{c}). The dominant GO term for each protein is
defined as its most frequent annotation within the respective
ontology. GFM Trained consistently places proteins with matching
dominant annotations in closer embedding neighborhoods than all
supervised baselines across all three ontologies and all values
of $k$.}
\label{fig:stringgo_crossonto}
\end{figure}


\begin{figure}[t]
\centering
\includegraphics[width=0.7\linewidth, page = 16]{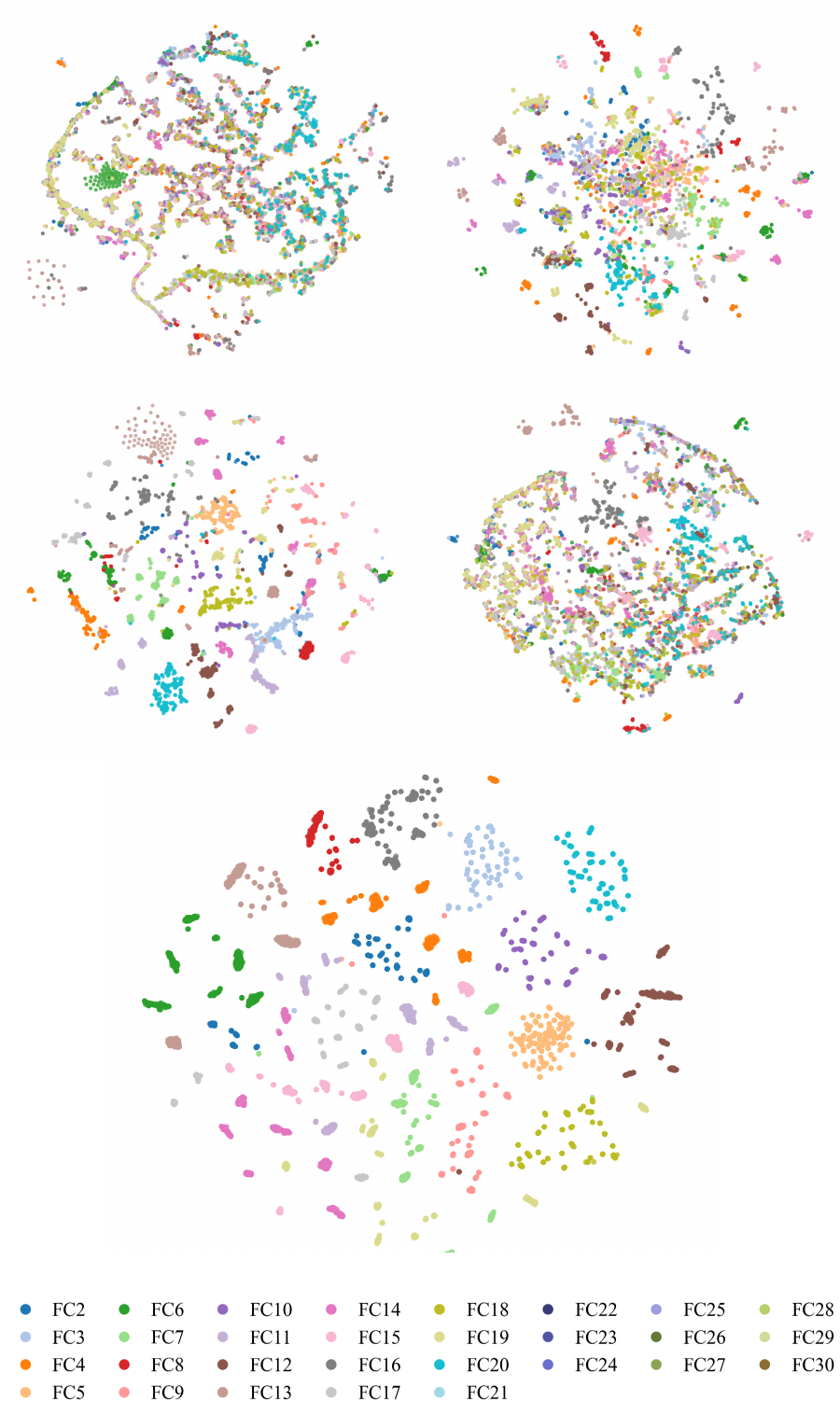}
\caption{\textbf{t-SNE visualization of protein embeddings across
SCOP fold classes in Fold-PPI.}
Each point represents a protein colored by its SCOP fold class
assignment (FC2--FC30). The four smaller panels (top two rows)
show embeddings from GFM zero-shot, GCN, GraphSAGE, and GAT.
The large bottom panel shows GFM Trained embeddings, in which
fold class clusters are visibly separated and compact. Each
cluster corresponds to one of the 29 labeled SCOP fold classes
present in the training split. The degree of intra-class
compactness and inter-class separation in GFM Trained embeddings
is substantially greater than in any baseline, indicating that
structural pretraining and fine-tuning produce a representation
space that organizes protein fold identity from interaction
network topology alone.}
\label{fig:foldppi_tsne}
\end{figure}


\begin{figure}[t]
\centering
\includegraphics[width=0.65\linewidth, page = 17]{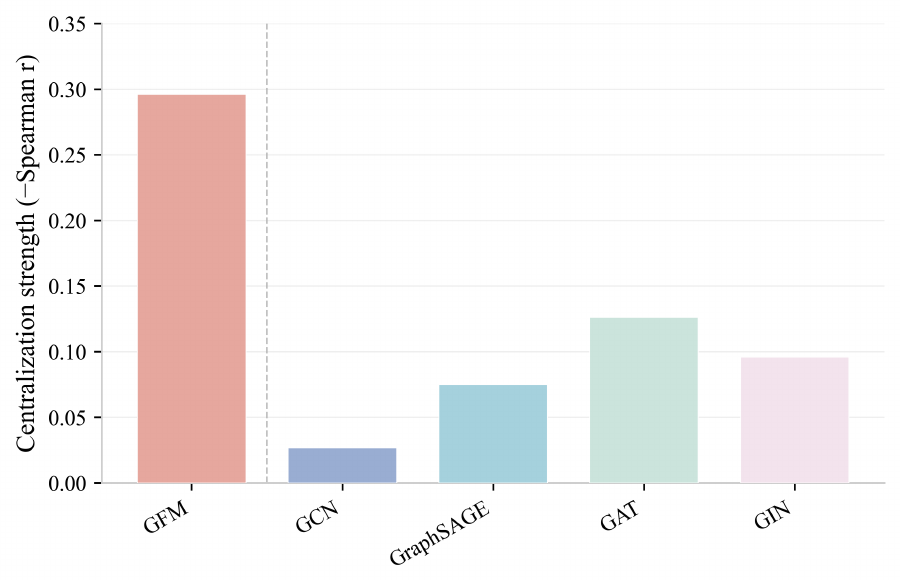}
\caption{\textbf{Tissue-prevalence centralization of fold class
embeddings in Fold-PPI.}
Bars show centralization strength, defined as the negative
Spearman rank correlation between the tissue prevalence of each
protein's fold class and its mean distance to the fold class
centroid in embedding space. A positive value indicates that
proteins belonging to fold classes that appear across more tissue
contexts are placed closer to their class centroid, reflecting
greater embedding compactness for broadly expressed structural
folds. GFM Trained produces a centralization strength of 0.296,
substantially exceeding all supervised baselines. GAT achieves
the highest baseline value of 0.126. The dashed vertical line
separates the GFM Trained model from supervised baselines.}
\label{fig:foldppi_central}
\end{figure}

\end{document}